\documentclass[journal]{IEEEtran}

\usepackage{graphicx,wrapfig,lipsum,subfigure}
\usepackage{tikz}
\usepackage{tikz-qtree}
\usetikzlibrary{trees} 
\usepackage{amsmath}
\usepackage{tikz}

\usepackage{caption} 
\usepackage{rotating}
\ifCLASSINFOpdf
\else
\fi
\begin{document}
\title{A Survey of Feature Types and Their Contributions for Camera Tampering Detection}

\author{Pranav~Mantini~and~Shishir~K.~Shah \\
Department~of~Computer~Science, 
University~of~Houston,~Houston,~TX-77204
}
\maketitle

\begin{abstract}
Camera tamper detection is the ability to detect unauthorized and unintentional alterations in surveillance cameras by analyzing the video. Camera tampering can occur due to natural events or it can be caused intentionally to disrupt surveillance. We cast tampering detection as a change detection problem, and perform a review of the existing literature with emphasis on feature types. We formulate tampering detection as a time series analysis problem, and design experiments to study the robustness and capability of various feature types. We compute ten features on real-world surveillance video and apply time series analysis to ascertain their predictability, and their capability to detect tampering. Finally, we quantify the performance of various time series models using each feature type to detect tampering. 

\end{abstract}
\section{Introduction}
\begin{figure}[htp]
	\centering
	\begin{tabular}{cccc}
		\includegraphics[width=.20\linewidth]{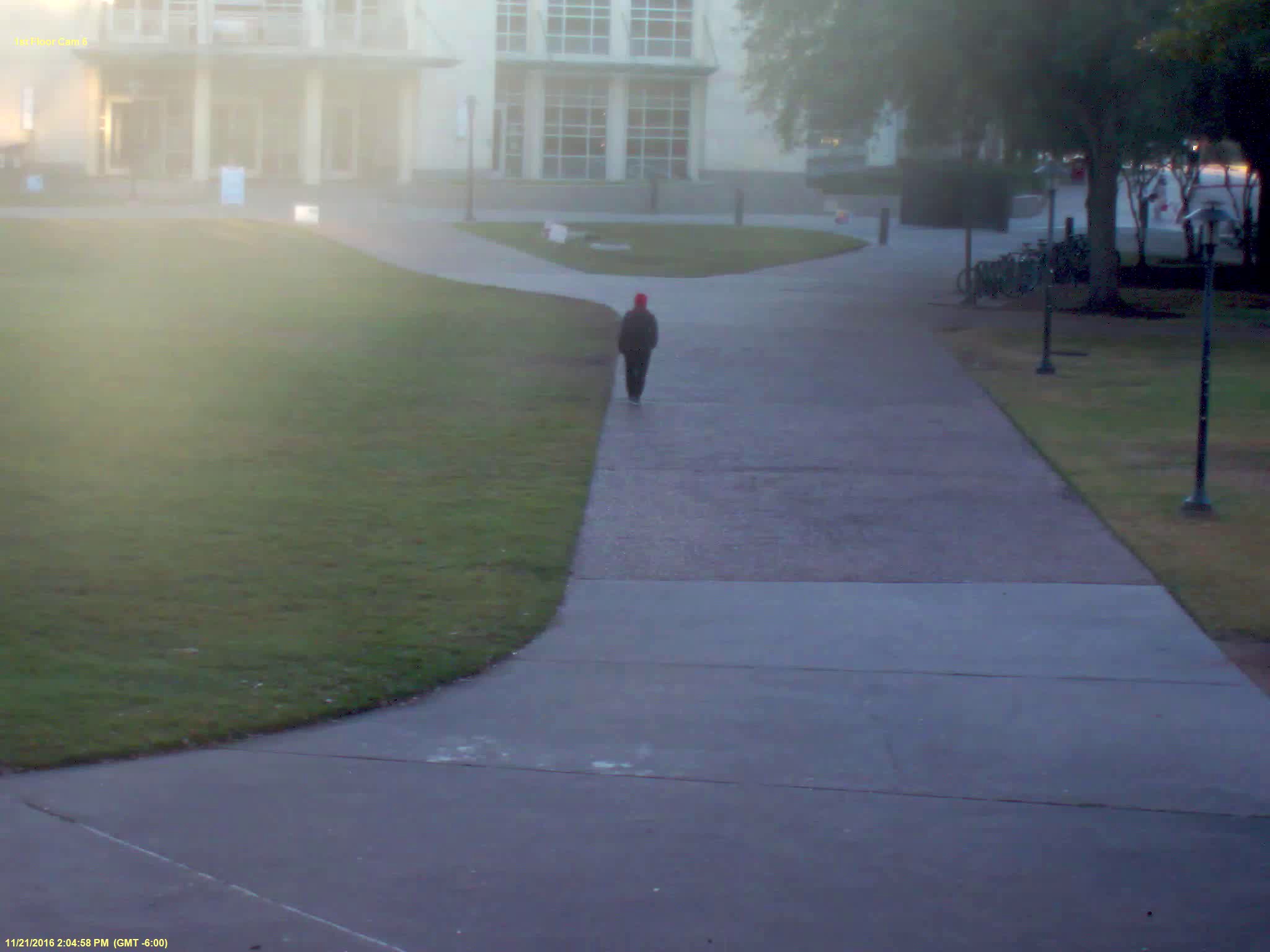}&
		\includegraphics[width=.20\linewidth]{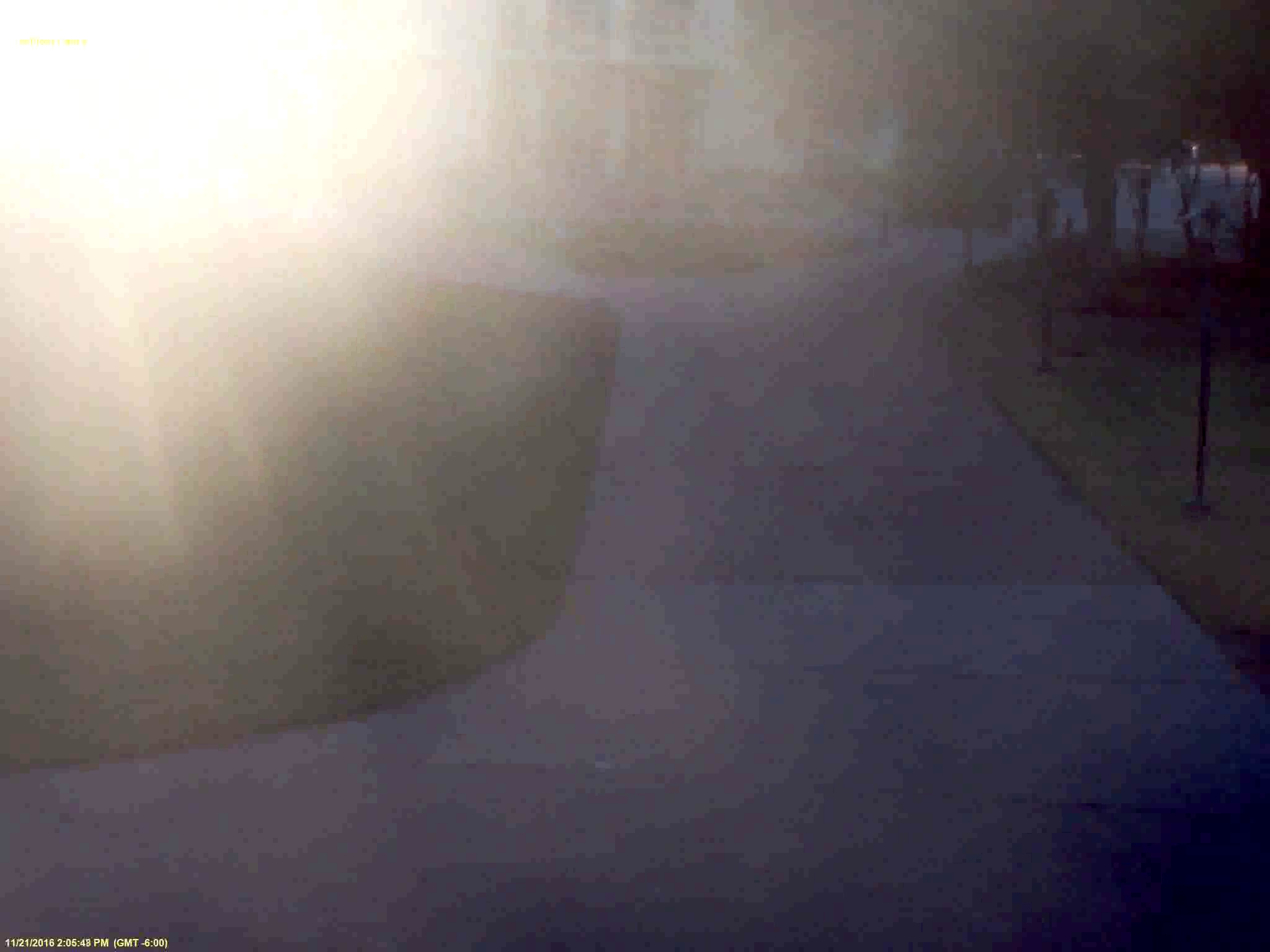}&
		\includegraphics[width=.20\linewidth]{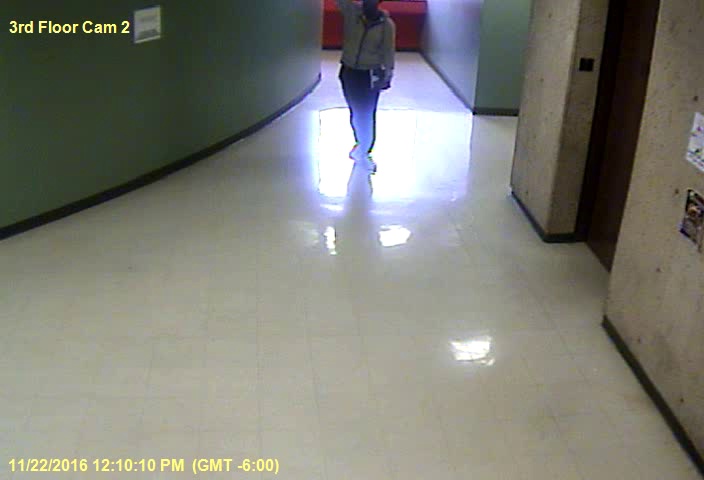}&
		\includegraphics[width=.20\linewidth]{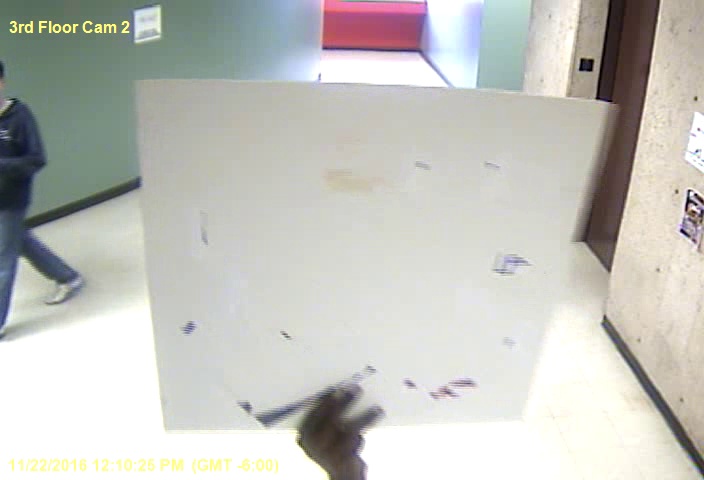}\\
		(a) Normal & (b) Tampered & (c) Normal & (d) Tampered\\
	\end{tabular}
	
	\caption{ \label{fig:natural_tamper} Example of natural tampering due to reflection of sunlight on to the camera lens (a) \& (b), Example of intentional tampering due to a covered lens (c) \& (d)}  
\end{figure}

\noindent An unauthorized alteration in the viewpoint of a surveillance camera is called tampering. This can occur due to natural phenomena, for example, the lens can accumulate dust, it can lose focus, the image/video quality can degrade, its viewpoint can shift, etc. Camera tampering can also be induced to accomplish malicious activities (like theft and property damage). Examples include spray painting, blocking, and/or changing the view of the camera. Though the repercussion of the latter event could be severe, necessitating immediate attention, the former event is detrimental as well for forensic needs. Figure~\ref{fig:natural_tamper} (b) shows an example of tampering due to a natural phenomenon, where the sunlight is reflected on to the camera lens for the scene observed in Figure~\ref{fig:natural_tamper} (a). Figure~\ref{fig:natural_tamper} (d) shows an example of tampering induced by a human for the scene observed in Figure~\ref{fig:natural_tamper} (c).

%

\vspace{-.25cm}
Techniques for automatically detecting tampering by analyzing the video are referred to as \textbf{camera tampering detection} techniques. These techniques are needed for ensuring the integrity of surveillance cameras and the efficacy of acquired video. This is especially the case for: 
\begin{itemize}
	\item \textbf{Large surveillance networks:} Surveillance cameras have become an integral part of public and private infrastructures in recent years. Today, large-scale surveillance systems are deployed frequently to enhance security. For example, a large university can have upwards of 1500 cameras, an international airport can have upwards of 3000 cameras, and a large casino can have well over 5000 cameras. Surveillance camera systems are deployed over a large area with a centralized control point. Such distribution amongst numerous electronic components demands rigorous maintenance through a continual review process. Currently, security officers periodically review cameras to ensure their functionality and identify camera tampering. Reviewing thousands of cameras manually to ensure functionality is a tedious task and is prone to human error. 
	
	\item \textbf{Enhanced security and reliability:}Videos captured by cameras are frequently used as forensic evidence. A tampered camera could result in the loss of valuable evidence.
	
	\item \textbf{Dependency of high-level algorithms:}  High-level computer vision algorithms like tracking~\cite{mantini2016multiple, yan2012track}, re-identification~\cite{mantini2015person, BedagkarGala2014270} and  motion prediction~\cite{mantini2014human} are designed assuming that the cameras are functioning properly resulting in tamper-free videos. Tampered and poor quality video from cameras lead to erroneous results in high-level algorithms.
\end{itemize}

\noindent In an attempt to address these needs, researchers have increasingly focused on camera tampering detection techniques over the past decade. 
Camera tampering detection is a challenging problem to solve due to:

\begin{itemize}
	\item \textbf{Illumination changes:}  Similar to many computer vision algorithms, camera tampering algorithms are affected by illumination changes in the environment. Outdoor cameras are exposed to varying illuminations and weather conditions. Illumination changes in indoor cameras may occur due to the switching off lights (on/off). The varying illumination and weather conditions often lead to high \textbf{false alarm} rates.	
	\item \textbf{Uncertainty in persistence of a tampering:} The temporal extent of tampering could be short-lived or persistent. For example, events involving large objects passing through the scene may block most of the camera view, but they are short-lived and not considered as tampering. Designing algorithms that are capable of distinguishing among these events is crucial. Camera tampering detection techniques should perform with an acceptable.
	 \textbf{detection rate}.
	\item \textbf{Limited training data:} Many computer vision algorithms learn specific features of a scenario through training data. However, cameras are deployed in a wide variety of scenarios from empty roads to crowded airports. It is not always possible to acquire a large amount of \textbf{training data} for each scenario.	
	\item \textbf{Limited compute resources:} These algorithms can be deployed either on-board the computer on the camera or at the central location where the videos are recorded/managed. In the earlier case, there are limited compute resources available, and in the latter case, the algorithm may be required to process large amounts of videos from multiple cameras. The \textbf{complexity} of the algorithms play a critical role in the performance of the system.	
\end{itemize}

\noindent In this paper, we formulate the problem of camera tampering detection as that of a change detection problem. We organize the existing work in this area based on the type of features used for detection and conduct a survey. Furthermore, we perform feature analysis to ascertain the capability of each feature under a common framework in real surveillance videos. We formulate the problem of camera tampering detection as a time
series analysis problem and perform an analysis of ten different features. Data from a surveillance camera is fit to ten different time series models based on each of the ten features. Prediction error in the model is used to study the behavior of the feature under normal operating conditions, and also quantify the ability of the feature to detect tampering. The contributions of this paper are:

\begin{itemize}
	\item We present a survey of the existing research organized by feature types.
	\item We present a novel formulation of tampering detection as a time series problem. 
	\item We perform feature analysis to ascertain the complexity, and ability to model various features.
	\item We present results to quantify the performance of various feature types to detect camera tampering.  
\end{itemize}

\section{Camera Tampering}
\noindent Most literature has classified camera tampering under three categories~\cite{Jimenez2007Automatic}: 

\begin{itemize}
	\item \textit{Covered tampering} occurs when the view of the camera is blocked, which results in an occluded image. This can happen when someone intentionally blocks the view of the camera. This can also occur due to natural factors like the accumulation of dust on the lens in outdoor cameras. Partial occlusions also may arise from foliage growing in front of the camera. 
	\item \textit{Defocussed tampering} occurs when the lens of the camera is not focused correctly, which results in a blurred video. This can happen when someone intentionally changes the focus of the camera or can occur due to natural factors, for example, day/night autofocus cameras change focus when switching between modes and sometimes fail to focus. 
	\item \textit{Moved tampering} occurs when the viewing direction of the camera is altered. This happens when someone intentionally changes the direction of the camera. This could also occur naturally due to strong winds.
	
\end{itemize}


%

\noindent Most researchers have developed individual algorithms to detect each tampering, while a few others have developed unified algorithms to detect the three types of tampering simultaneously. Unified tampering detection algorithms have low complexity at the expense of their inability to classify the tamperings. Independent tamper detection algorithms are designed for each individual tamper but may have common preprocessing and training stages. The choice of one over the other is driven by the application and the computational resources available. 

\section{Camera Tampering Detection Techniques}
\noindent 
More recently, with the ubiquitous deployment of low-cost cameras for surveillance, more research has been dedicated towards robust automatic detection of camera tampering, leading to a handful of patents~\cite{flook2007camera, skans2011camera}.  While there might be considerable overlap between published research in this area and patents, this work focuses on summarizing the published research. As shown in Figure~\ref{fig:struct}, the overall approach for camera tampering detection consist of feature extraction, reference model design, and detection mechanism. Feature extraction is a crucial step that determines, the capability of the system to distinguish between tampered and normal images, as a result, a number of features have been explored in the past. 

In this section, we organize and survey existing methods based on the type of features used for tampering detection. In section~\ref{sec:IV} we present a time series analysis of tampering detection techniques based on feature types. \vspace{-.3cm}

\subsection{Taxonomies of Camera Tampering Detection}
\begin{figure}[ht]
	\centering
	\includegraphics[width=0.9\linewidth]{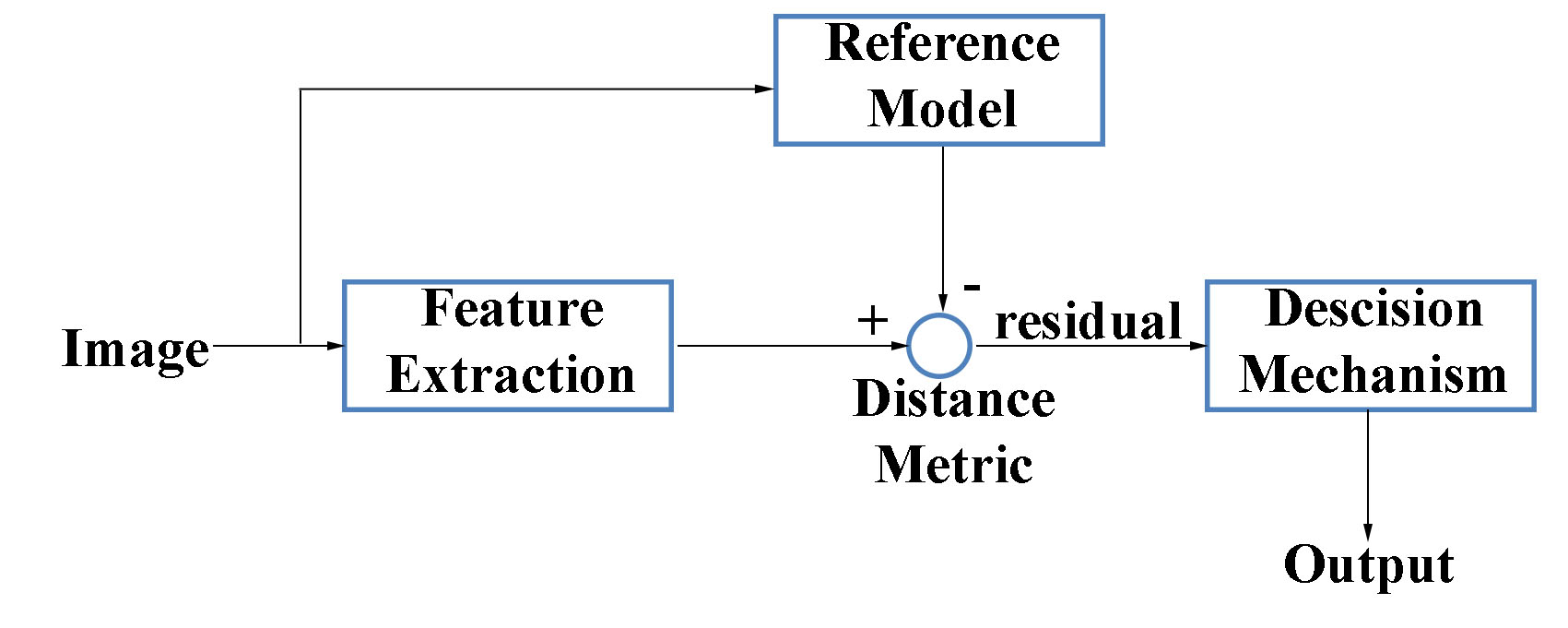}
	\caption{Residual based structure for camera tampering detection\label{fig:struct}}
\end{figure}

\noindent Camera tampering detection can be regarded as a sub-problem of video change detection~\cite{meng1995scene}, where the objective is to detect changes in the scene. Willsky \textit{et al.}~\cite{willsky1985detection} proposed a residual-based structure for detecting abrupt changes in dynamic systems. The majority of methods in camera tampering detection can be described using a similar structure, as is depicted in Figure ~\ref{fig:struct}.

The model assumes that certain features of the image remain consistent under normal operating conditions. The feature extraction step ($T$) takes test images $I_t$ as input and computes a set of features. The reference model $F$ predicts a reference set of features that represents the normal operation of the camera. Often times there exists a set of reference images $I_r$ that represent normal operating conditions of the camera, which the reference model uses to compute the reference features. The computed features are compared against the reference features using a distance metric ($D$). The decision mechanism detects tampering based on the amount of residual $R$, (distance between the reference and computed features). The residual can be computed as:

\begin{equation}
R = D(T(I_t), F(I_r))
\end{equation}

Owing to the dynamic nature of the images captured by surveillance cameras, often times the reference images ($I_r$) are gradually replaced with test images ($I_t$) to enable an online learning system. For example, this can allow the reference model to adapt to naturally occurring illumination changes. The feature extraction consists of a set of image processing ($IP$) algorithms applied in series to the test images over which statistical measure (S) is computed. The reference model applies the same set of transformations and statistical measures on the reference images.

\begin{equation}
\begin{split}
T(I_t) = S(IP_2(IP_1(I_t))) \\
F(I_r) = S(IP_2(IP_1((I_r)))
\end{split}
\end{equation}

Taking inspiration from research in Image Quality Assessment~\cite{wang2006modern}, Wang \textit{et al.}~\cite{Wang2011Real} classified tampering methods as full-reference, reduced-reference, and no-reference techniques. This is a well-known taxonomy based on the characteristics of the types of features. 

\textbf{Full reference:} The feature used is an image. A pixel-wise comparison is conducted to assess the residual, which is used for detecting tampering. An example of a full-reference technique is where we assume that majority of the edges in images remain approximately the same unless there is tampering. In this case, the feature extraction step computes the edges of the input image. The output from the reference model is the expected edges under normal conditions. The intersection between two sets of edges is used to compute a residual value. Let $E(I)$ be an edge detection operation that outputs edges as a binary image of the input image $I$. Let $I_t$ and $I_r$ be the test image and the reference image, respectively. Then the residual is computed as:

\begin{equation}
R = \cap(E(I_t), E(I_r)), \text{where $\cap$ is the intersection}
\label{eq:endge_intersection_residual}
\end{equation}

One can employ thresholding on the count of the residual edges as a decision mechanism for tampering detection~\cite{Lin2012Real}.

\textbf{Reduced reference:} In contrast to full-reference methods, reduced reference approaches compute a statistical or numerical quantity from the image pixels or a subset of the image pixels to represent the image. The residual is obtained by a comparison of features computed from the test and reference images. A pixel-wise comparison is not conducted in the reduced reference techniques. An example of a reduced reference technique is, where we assume that the entropy of the background pixels in an image stays approximately constant unless there is tampering. In this case, the feature extraction step computes the entropy of the background pixels in the input image. The output from the reference model is the expected entropy of the background pixels when the camera is functioning under normal conditions. Let $B(I)$ be the background, and $\epsilon(I)$ be the entropy of an $I$. Then a residual is computed as follows.

\begin{equation}
R = \setminus(\epsilon(B(I_t)), \epsilon(B(I_r))),  \text{where $\setminus$ is the difference}
\label{eq:entropy_bg_residual}
\end{equation}
One can again employ thresholding on the residual as a decision mechanism for tampering detection~\cite{Ellwart2012Camera}. 

\textbf{No-reference:} The prior techniques require training data (normal images) to create a reference model based on normal operations. In contrast, the no-reference techniques hypothesize a reference model. An example of the no-reference technique is where we assume that if a minimum of $x\%$ of the pixels in an image is black, then covered tampering has occurred. In this case, the output from the image processing algorithm could be a value representing the percentage of pixels that are black. This value is compared to the value hypothesized as a reference. Hence, if $IP_b$ is an algorithm that counts the number of black pixels, then the residual can be computed as

\begin{equation}
R = sign(\setminus(IP_b(I_t), x)), 
\end{equation}
where the sign is the signum function that returns $+1$ for positive, and $-1$ for negative numbers.

No-reference methods have lower computational overhead compared to full-reference methods. Nonetheless, they are difficult to model to account for the range of expected situations leading to camera tampers. Reduced-reference can be a feasible compromise among the three types of approaches~\cite{Wang2011Real}.


\noindent Feature extraction is a crucial step that determines the capability of an approach to distinguish between tampered and normal images. As a result, a number of features have been explored in the past. Also, a study of the various features under normal operating conditions is vital in understanding the performance of the algorithms, especially because they influence the false alarm rate. In this paper, 
first we organize and review the various features used for camper tampering detection. Second, we conduct a study to understand the effect of computed features over the performance of the tampering detection algorithms. 
\vspace{-.3cm}

%
%
%
\subsection{Feature Extraction}
The framework for camera tampering detection techniques bet on the idea that some property of the image remains consistent under normal operating conditions, even if it is for a short period of time. Some researchers have worked with the assumption that it is the background that remains consistent, while others have assumed that it is the edges, and yet a few that assume it is the interest points. Camera tampering detection techniques can be categorized as background modeling, edge modeling, and interest point modeling techniques.

\noindent \textbf{Background modeling techniques:} Background refers to the elements of the scene that do not undergo motion. Pixels that belong to the background do not undergo an abrupt change in appearance unless subjected to sudden global illumination changes (eg. lighting in outdoor scenarios and a light turned off/on in indoor scenarios). Many methods have leveraged this idea to model background as a feature for tampering detection. Some methods have employed an efficient frame differencing model~\cite{collins2000system,Aksay2007Camera,Saglam2009Real,Kryjak2012FPGA,Guler2016Real} to compute the background, while others have modeled the background as a mixture of Gaussians~\cite{czyzewski2008moving,Ellwart2012Camera}, and a few have modeled the background using  codebooks~\cite{Tung2012Camera,kim2005real}.

Following the residual-based structure proposed by Willsky \textit{et al.}~\cite{willsky1985detection}, we can conduct a pixel-wise comparison between the reference background and the test background image to compute a residual. Saglam \textit{et al.}~\cite{Saglam2009Real, Guler2016Real} modeled two backgrounds separated by a time delay and conducted a pixel-wise comparison between the backgrounds to compute a residual. A thresholding based detection mechanism is employed to detect moved tampering. Huang \textit{et al.}~\cite{Huang2014Rapid} used the absolute difference between the time-delay backgrounds to detect moved tampering and as an initial test for detecting covered tampering. Block matching algorithm uses a pixel-wise error computation to detect shifts between images. Ellwart \textit{et al.}~\cite{Ellwart2012Camera} employed a modified block matching algorithm (a shift detection algorithm based on block matching) to compute a residual, and a thresholding mechanism to detect moved tampering.  Most methods have preferred to apply a statistical measure on the background images, and compare the measures to arrive at a residual.

Entropy, defined as: 
\begin{equation}
E = - \sum_{k}{p_k . \ln(p_k)}, 
\end{equation}
where $p_k$ is the probability of pixel k in the image, is a statistical measure of randomness in the image.  Ellwart \textit{et al.}~\cite{Ellwart2012Camera} used the difference between the entropy of the reference and test background as a residual (see Eq.~\ref{eq:entropy_bg_residual}) and employed thresholding on this residual as a decision mechanism to detect covered tampering. On the other hand, some methods have used the histograms of the reference and test background to compute a residual. A common observation amongst researchers is that covering a camera would result in the concentration of histogram within a small range of intensity values. For example, if the lens of a camera is covered, most of the pixels would be close to black and the histogram of the image would be concentrated around the lower intensity values. The count of the histogram values around the maximum was used as a measure and compared it with the histogram of the test image to detect a covered tampering~\cite{Aksay2007Camera,Saglam2009Real,Huang2014Rapid, Guler2016Real}. Lin and Wu~\cite{Lin2012Real} also extended such histogram analysis to detect defocussed and moved tampering. Let $H(I)$ be the histogram of an image $I$, $max(H(I))$ be the value of the histogram bin that has a maximum value. Let $K$ be the neighborhood around the maximum bin that we would like to accumulate values of the histogram, and let $B$ be the expected background for $I$. Then the residual can be computed using 

\begin{equation}
R = \setminus(\sum_{a = max(H(I))-K}^{max(H(I))+K}H_a(I),\sum_{a = max(H(I))-K}^{max(H(I))+K}H_a(B))
\end{equation}

Tung \textit{et al.}~\cite{Tung2012Camera} trained an adaptive background codebook model for classifying foreground and background pixels. The length of the codebook was used as a feature. The difference in the lengths between the background codebook and the test image codebook is used to compute a threshold.

\noindent \textbf{Edge modeling techniques:}
Edges in a scene correspond to discontinuities in-depth, surface orientations, material properties, and variation in scene illumination~\cite{barrow1981interpreting}. Modeling edges for camera tampering detection assumes that these properties remain unchanged under normal operating conditions. While this could be a good assumption for the first three properties, it is not so for the fourth property. Assuming scene illumination does not change could be detrimental for designing camera tampering detection algorithms. However, most tampering detection algorithms use an adaptive update scheme to model the reference features, which compensates for illumination changes. Furthermore, edges are more robust towards global illumination changes compared to the background. Edges correspond to sharp intensity changes in an image. A plethora of techniques is available to identify edges. Some methods have computed a simple pixel-wise gradient~\cite{mantini2017signal, Harasse2004Automated, Tsesmelis2013Tamper} to detect sharp intensity changes or make use of image processing filters like Sobel~\cite{Gaibotti2015Tampering, Lee2015unified,Huang2014Rapid,Shih2013Real,Raghavan2012Detection, Ellwart2012Camera, Ribnick2006Real} and Prewitt. Some have employed more sophisticated edge detection methods like a Canny edge detector~\cite{Lee2014Low,Wang2014Traffic,Lin2012Real, Ellwart2012Camera, Wang2011Real,Sidnev2018Efficient} to reduce noisy edges. On the other hand, a similar effect on images can be obtained by performing filtering in the frequency domain. For example, transforming an image into the frequency domain and applying a high pass filter followed by an inverse transformation can isolate the high-frequency content in the image~\cite{Guler2016Real, Saglam2009Real, Huang2014Rapid}.

Following the residual-based structure proposed by Willsky \textit{et al.}~\cite{willsky1985detection} to detect tampering, one can compute a pixel-wise comparison between the reference edge image and test edge image to arrive at a residual. Covered tampering could result in the disappearance of edges that are present in the reference image. Lin and Wu~\cite{Lin2012Real} considered the intersection of the edges detected between the reference and test images, this produced a binary image that represents the intersection of the edges, and the count of the intersecting edges was used as a residual (see Eq~\ref{eq:endge_intersection_residual}). This residual was thresholded to detect covered tampering, and as well as defocussed and moved tampering. A shift in the edges could correspond to moved tampering. Some methods have employed a block matching algorithm to compute the translation in edges between the reference and test image. A large value of translation parameters indicated moved tampering. Thresholding of the translation parameters was used as a detection mechanism to detect moved tampering~\cite{Harasse2004Automated,Jimenez2007Automatic}.

Defocussing results in a decrease in the sharpness of the image resulting in a degradation of edge content as well. Most methods have chosen to apply a statistical measure on the edge content to detect defocussed tampering. A general approach to edge detection is to compute a gradient and apply thresholding to determine the edges. Some methods assume the gradient magnitude to remain unchanged under normal operating conditions and applied a first-order or a second-order gradient (in the horizontal and vertical direction) to compute the gradient magnitude at each pixel, which represents the strength of the edge at each pixel. The sum or mean of the gradient magnitudes was used as a feature for tampering detection. Gaibotti \textit{et al.}~\cite{Gaibotti2015Tampering} used the difference of these features between the reference and test image to compute a residual for defocussed tampering detection. Mantini and Shah~\cite{mantini2017signal} used a similar approach for their unified tampering detection method. On the other hand, we can threshold the gradient magnitude to only retain the relevant edges, and accumulate the gradient magnitude at these locations to arrive at a feature value~\cite{Ellwart2012Camera,Harasse2004Automated}. The feature value obtained from the latter could be less noisy than accumulating the gradient magnitude from the entire image. Mantini and Shah~\cite{mantini2014human} compensated for this by applying a statistical filtering process on the computed feature values. One can also use a count of the number of edges as a feature for tampering detection. The difference in the count of the edges between the reference and test images can be used as a residual. Jimenez \textit{et al.}~\cite{Jimenez2007Automatic}, Wang \textit{et al.}~\cite{Wang2011Real}, and Huang \textit{et al.}~\cite{Huang2014Rapid} used thresholding as a detection mechanism on this feature for detecting defocussed tampering detection. Shih \textit{et al.}~\cite{Shih2013Real} followed a similar approach in their unified tampering detection method. Some methods have applied such an analysis in the frequency domain as well. Edges correspond to the high-frequency content in the image. A frequency transformation technique is applied, and high-frequency content of the image is accumulated in the frequency domain, this is used as a feature for tampering detection. Saglam \textit{et al.}~\cite{Saglam2009Real} and Aksay \textit{et al.}~\cite{Aksay2007Camera} applied wavelet transform on the image and used the sum of the coefficients corresponding to the high-frequency content as a feature for detecting defocussed tampering. Huang \textit{et al.}~\cite{Huang2014Rapid}, and Guller \textit{et al.}~\cite{Guler2016Real} followed a similar approach by applying discrete Fourier transform on the images.

A statistical measure such as entropy can also be applied on the edges to quantify the information in the image. The residual is obtained by taking the difference between the entropy of the reference with the test image. Harasse \textit{et al.}~\cite{Harasse2004Automated}, and Jimenez \textit{et al.}~\cite{Jimenez2007Automatic} used thresholding as a detection mechanism on this residual value to detect covered tampering. Computing entropy on the edges could allow for a more robust feature. Furthermore, statistical measures such as the sum and the mean of features could be more prone to noise.  Most of these methods, model the magnitude of the edge as an invariant feature, but one can consider the orientation of the edges as well. Modeling edge orientation could provide a robust model towards illumination changes compared to just the magnitude~\cite{dalal2006human}. Ribnick \textit{et al.}~\cite{Ribnick2006Real} considered the histogram of the orientation of the edge pixels as a feature to represent the reference and test image. The sum of the absolute difference between the histograms is used as a residual, and thresholded to detect tampering.

\noindent\textbf{Background and edge combined modeling techniques:} Some methods proposed a combination of background and edge detection to compute robust features. Saglam \textit{et. al}~\cite{Saglam2009Real} extracted the high-frequency content of the background image and used it as a reference for defocussed detection. Lee \textit{et al.}~\cite{Lee2014Low, Lee2015unified} applied edge detection on the background image to create a reference for unified tampering detection. The intersection between the reference and test image was used to compute a residual and used thresholding as a detection mechanism for detecting tampering.

\noindent \textbf{Interest point modeling techniques:} 
Lindberg~\cite{lindeberg2013scale} described an interest-point as one that has a mathematically well-founded definition, a well-defined position in image space, rich local information content in the surrounding, and stability under local and global perturbations in the image domain such as illumination/brightness variations. This definition allows the interest points to be reliably computed with a high degree of repeatability. Modeling interest points for camera tampering detection assumes that these interest points remain consistent under normal operating conditions of the camera. Researchers have used corner detection methods to model such features for camera tampering detection. The driving idea is to extract key points using Scale Invariant Feature Transform (SIFT)~\cite{lowe1999object, Yin2013Sift, Tsesmelis2013Tamper} from the reference images and use these interest points as a representation. Some methods have used Speeded-Up Robust Feature (SURF)~\cite{bay2006surf} as well, which are a computationally efficient version for computing SIFT~\cite{Javadi2015Video}

Commonly various statistical measures are applied to obtain features from the SIFT points. A count of the number of interest points is an example. Tsesmelis \textit{et al.}~\cite{Tsesmelis2013Tamper} used the difference in counts between the reference and test image as a residual. Yin \textit{et al.}~\cite{Yin2013Sift} used a SIFT based image described in Equation~\ref{eq:SIFT_image_desc} as a feature. 

\begin{equation}
F(I) = \frac{1}{n} \sum_{i=0}^{n}\sqrt{x_i^2+y_i^2} | D(a_i)|
\label{eq:SIFT_image_desc}
\end{equation}

\noindent where $n$ is the number of SIFT key-points from the input image, $D(a)$ is the response value for the key points $a$; $x$ and $y$ is the vertical and horizontal coordinates of the key points. The difference in the SIFT based descriptors between the reference and test image is used as a residual and thresholded to detect covered and moved tampering. Javadi \textit{et al.}~\cite{Javadi2015Video} performed matching of interest points between the reference and test image to detect tampering, and matched SIFT points between the reference image and test image to estimate the global motion. The displacement obtained from the homography computation was used as a residual. A thresholding mechanism on this value was used to detect tampering.\\
\textbf{Deep learned features:} Recently, researches have found success in training Convolutional Neural Networks (CNNs) to detect visibility loss~\cite{Ivanov2019Visibility}, and detecting color and intensity based abnormality~\cite{Dong2016Camera}. Mantini and Shah~\cite{mantini19camera} proposed a deep learning approach with extended ability to detect covered, defocussed, and moved tampering. The reference model consists of a deconvolutional neural network that learns the probability distribution of normal images, and samples from it to generate reference images. Then the test and reference images are transformed to a new features space and classified as tampered or normal images.

\vspace{-.3cm}

\subsection{Reference Model}
The reference model generates the expected features under normal operating conditions. Residual is computed by comparing this against the features of the test image. The input to reference model is usually a set of images that represent the camera under normal operating conditions. This data is not available. A general strategy is to assume temporal constancy, where frames from the immediate past are used as reference images. A common technique is to use a combination of the reference images to arrive at a reference value. This allows the system to adapt to naturally occurring illumination changes. For example, Jimenez et al.~\cite{Jimenez2007Automatic,Ellwart2012Camera,Aksay2007Camera,Saglam2009Real,Huang2014Rapid} updated the background reference image using a moving average model, and Wang et al.~\cite{Wang2011Real,Lin2012Real,Ellwart2012Camera,Harasse2004Automated} accumulated the edges over a set of frames to form reference edges. Mantini and Shah~\cite{mantini2019uhctd}, have trained and extracting features from various scene classification neural networks, and used them for classifying images under various tampering and normal class.

Assuming temporal constancy has disadvantages. If images in the immediate past have tampered, then the model accumulates these features as well. As a result, the reference model drifts and the approach fails to detect tampering. Adversely, the system falsely identifies normal frames as tampered (false positives). Selectivity is a common technique to avoid this, where frames identified as normal are selectively included in the reference model. However, the performance of the system is contingent on its ability to detect tampering.  \\

\noindent\textbf{Detection Mechanism}\\
The detection mechanism analyzes the distance between features of the reference image and test image and labels the image as either tampered or normal. It takes as input a residual value and maps it to a decision. A linear decision boundary using a thresholding scheme has been the norm~\cite{Saglam2009Real,Jimenez2007Automatic,Wang2011Real,Lin2012Real,Ellwart2012Camera,Aksay2007Camera}. Some methods have proposed multiple thresholds~\cite{Huang2014Rapid}. Lee \textit{et al.}~\cite{Lee2014Low} proposed an adaptive threshold, producing a non-linear boundary to cope with the complexity. However, a thresholding mechanism has limitations and parameter tuning is often required to choose an appropriate threshold. A non-linear decision making capability may cope better with the complexity of observations from surveillance cameras.

\section{Analysis of Feature Type for Camera Tampering Detection Techniques}
\label{sec:IV}
There is a certain amount of similarity amongst the previous approaches, in how the reference modeling is formed, and the detection mechanism is applied to detect tampering. However, the methods vary in their choice of feature type that they assume to be the most discriminative between normal and tampered images. Most methods choose a threshold on the residual to detect tampering. The detection mechanism models the value of the residual by placing an assumption on its behavior - it assumes that the value of the residual does not exceed the bounds defined by the threshold - under normal operating conditions of the camera. The choice of the feature dictates the behavior of the residual. For example, global illumination change affects the background of the image more than the edges, a residual computed from the background can vary largely compared to a residual computed from the edges. As part of this analysis, we study the behavior of the residual to ascertain the robustness of different feature types used for camera tampering detection.

First, we choose ten features for analysis. Then we cast the problem of camera tampering detection as a time series analysis problem. Data from a surveillance camera is fit to ten different time series models based on each of the ten features. We compute prediction error in the model to study the behavior of the feature under normal operating conditions, and finally quantify the ability of the feature to detect tampering as well. 
\vspace{-0.3cm}
\subsection{Feature Selection for Analysis}
We choose ten features of which four leverage background, four leverage edges, and two leverage key-points as features to study:
\begin{enumerate}
	\item Sum of foreground pixels ($b_1$): We compute this as a baseline to study the relationship between the background and foreground pixels in the video. The hypothesis is that the number of foreground pixels in the images under normal operating conditions can be modeled. Let $B_t$ be the background of the image $I_t$ at time $t$. The residual at time $t$ is given by:
	\begin{equation}
	r^{b_1}_{t} = \sum_{(i,j) \in I} (B_t(i,j) - I_t(i,j))
	\end{equation}
	\item Difference in entropy of background and foreground ($b_2$): This residual is similar to the one proposed by Ellwart \textit{et. al}~\cite{Ellwart2012Camera} to detect covered tampering. The hypothesis is that the difference in the entropy of the background and entropy of the image can be modeled. Let $\epsilon(B_t)$, and $\epsilon(I_t)$ be the entropy of the background $B_t$, and the image $I_t$ at time $t$. The residual at time $t$ is given by:
	\begin{equation}
	r^{b_2}_{t} = \epsilon(B_t) - \epsilon(I_t)
	\end{equation}
	\item Difference in time delayed backgrounds ($b_3$): This feature is similar to the one proposed by Saglam \textit{et al.}~\cite{Saglam2009Real}, and Huang \textit{et al.}~\cite{Huang2014Rapid} to detect moved, and covered tampering, respectively. The hypothesis is that the change in the background between a small predefined period of time can be modeled. Let $B_t$, and $B_{t-n}$ be the background of images at time $t$, and $t-n$, respectively. The residual at time $t$ is given by:
	\begin{equation}
	r^{b_3}_t = \sum_{(i,j) \in I} (B_t(i,j) - B_{t-n}(i,j))
	\end{equation}
	\item Difference in maximum values of the histogram ($b_4$): This feature is similar to the one used in~\cite{Aksay2007Camera,Saglam2009Real,Guler2016Real,Huang2014Rapid} to detect covered tampering. The hypothesis is that the count of the pixels concentrated around the maximum in the histogram of the image can be modeled. 
	Let $H_{I_t}$ and $H_{B_t}$ be the histogram of the image, and the background at time $t$. Then the residual at time $t$ is given by:
	
	\begin{equation}
		r^{b_4}_t = \sum_{a = m_{I_t} - K}^{m_{I_t} + K} H_{B_t}(a) - \sum_{a = m_{B_t} - K}^{m_{B_t} + K} H_{I_t}(a)
	\end{equation}  
	where $m_{I_t} =   \arg {max(H_{I_t}(h))}$ and $m_{B_t} = \arg {max(H_{B_t}(h))}$, and $K$ is a constant.
	\item Difference in count of edges ($e_1$): This is a baseline metric to understand the behavior of edges in the video. The hypothesis is that the count of the edges in each image of the video can be modeled.
	Let $\bar{E}_t$ be the reference edges, obtained by accumulating edges, which is a moving average of the edges from the previous frames.

	\begin{equation}
	\label{EQ:Bg_moving_average}
	\bar{E}_t = \alpha \bar{E}_{t-1} + (1-\alpha) E_{t-1}
	\end{equation}  
	where $E_t$ are the edges at time $t$.  Let $C(\bar{E}_t)$ and $C(E_t)$ be the count of the number of edges in the reference, and the test image at time $t$. The residual at time $t$ is given by:
	\begin{equation}
	r^{e_1}_t = C(\bar{E}_t) - C(E_t)
	\end{equation} 
	
	\item Difference in gradient magnitude of edges ($e_2$): This feature is similar to the one proposed by Gaibotti \textit{et al.}~\cite{Gaibotti2015Tampering} to detect defocussed tampering. The hypothesis is that the sum of gradient magnitude in each frame of the video can be modeled. Let $\bar{E_t}$ be the reference edges obtained by accumulating the gradient magnitude over the previous frames using a moving average (Eq~\ref{EQ:Bg_moving_average}). Let $E_t$ be the gradient magnitude of the edge at time $t$. Then the residual is given by:
	\begin{equation}
	r^{e_2}_t = \sum_{(i,j) \in I} (\bar{E}_t(i,j) - E_t(i,j))
	\end{equation}
	\item Difference in gradient magnitude of strong edges ($e_3$): This feature is similar to $e_2$, with the exception that the gradient magnitude is thresholded to retain strong edges. This is similar to the feature used by Ellwart \textit{et al.}~\cite{Ellwart2012Camera}, and Akshay \textit{et al.}~\cite{Aksay2007Camera}.
	\item Count of intersecting edges ($e_4$): This feature is similar to the one proposed by Lin and Wu~\cite{Lin2012Real} to detect defocussed and moved tampering. The hypothesis is that the location and count of edges can be modeled under normal operating conditions of a camera. Let $\bar{E}_t$ be the reference edges as defined in Equation~\ref{EQ:Bg_moving_average}. The residual at time $t$ is given by:
		\begin{equation}
	r^{e_4}_t = \sum_{(i,j)\in I} 1(\bar{E}_t(i,j), E_t(i,j))
	\end{equation}
	where 
	\[   
	1(a,b) = 
	\begin{cases}
	\text{1,} &\quad\text{a = b}\\
	\text{0,} &\quad\text{otherwise.} \\ 
	\end{cases}
	\]
	$\bar{E}_t$ and $E_t$ take binary values, representing either the presence or absence of an edge.
	
	\item Difference in count of keypoints ($k_1$): This is a baseline feature we compute to understand the behavior of the interest points in the video. The hypothesis is that the count of keypoints in each frame of the video can be modeled. The background is used as a reference image, and the key-points computed on the background image are used as a reference. Let $C(\bar{K}_t)$, and $C(K_t)$ be the count of key-points in the reference and the current image  at time $t$. The residual is given by:
	\begin{equation}
	r^{k_1}_t = C(\bar{K}_t) - C(K_t)
	\end{equation}
	\item Difference in keypoint descriptors ($k_2$): This feature is similar to the one computed by Yin \textit{et al.}~\cite{yan2012track} to detect covered, and moved tampering. The hypothesis is that the key-point descriptor (Equation~\ref{eq:SIFT_image_desc}) computed on each image can be modeled. Let $F(\bar{K}_t)$ and $F(K_t)$ be the descriptor computed on the background and the input image at time $t$. The residual is given by:
	\begin{equation}
	r^{k_1}_t = F(\bar{K}_t) - F(K_t)
	\end{equation}
	
\end{enumerate}

\subsection{Problem Formulation}
Let $\{R^{f_1}, R^{f_2}, .... , R^{f_n}\}$ be residual series generated by feature $f_i \in \{b_1, b_2, b_3, b_4, e_1, e_2, e_3, e_4, k_1, k_2\}$, such that $R^{f_i} = \{r^{f_i}_t, r^{f_i}_{t-1}, ..., r^{f_i}_1\}$, where $r^{f_i}_t$ is the residual computed using feature $f_i$ at time $t$.\\

\noindent\textbf{Residual series as a $ARIMA(p,d,q)$ process}

The residual $r^{f}_t$ computed from feature $f$ at time $t$ is often represented as the difference between the reference feature $\bar{f}_t$ computed at time $t$ and the feature $f$ computed at time $t$. We represent $r^f_t$ as a linear combination of the $\bar{f}_t$ and $f$.
\begin{equation}
\label{eq:general_residual}
r^f_t = \alpha \bar{f}_t + \beta f_t
\end{equation}
The reference feature is often computed as a moving average of the features computed over the previous time instants. Akshay \textit{et al.}~\cite{Aksay2007Camera} and others~\cite{Saglam2009Real,collins2000system,Kryjak2012FPGA,Guler2016Real} have employed a moving average of the previous backgrounds to compute the reference background at time $t$. In equation~\ref{EQ:Bg_moving_average}, we model the reference edges at time $t$ as moving average of edges computed over the previous time instants. We generalize this and represent the reference feature $\bar{f}$ as a linear combination of reference feature computed over previous time instants.
\begin{equation}
\label{eq:general_ref_feature}
\bar{f}_t = \gamma \bar{f}_{t-1} + \delta f_{t-1}
\end{equation}
From equations~\ref{eq:general_residual} and \ref{eq:general_ref_feature}, we have
\begin{equation}
\label{eq:general_ref_feature_2}
\begin{split}
r^f_t &= \alpha(\gamma \bar{f}_{t-1} + \delta f_{t-1}) + \beta f_t\\
	  &= \alpha\gamma \bar{f}_{t-1} + \beta f_t + \alpha\delta f_{t-1}
\end{split}
\end{equation}
From equation~\ref{eq:general_residual}, we have $\bar{f}_{t-1} = \frac{1}{\alpha}[r^f_{t-1} - \beta f_{t-1}]$, substituting in Equation~\ref{eq:general_ref_feature_2},
\begin{equation}
\label{eq:residual_t}
\begin{split}
r^f_t &= \alpha\gamma [\frac{r^f_{t-1} - \beta f_{t-1}}{\alpha}] + \beta f_t + \alpha\delta f_{t-1}\\
r^f_t &= \gamma r^f_{t-1} + \beta f_t + (\alpha\delta - \beta\gamma) f_{t-1}
\end{split}
\end{equation}
Previous residual can be written using equation~\ref{eq:residual_t} as:
\begin{equation}
\begin{split}
r^f_t - \gamma r^f_{t-1} &=  \beta f_t + (\alpha\delta - \beta\gamma) f_{t-1}\\
r^f_{t-1} - \gamma r^f_{t-2} &=  \beta f_{t-1} + (\alpha\delta - \beta\gamma) f_{t-2}\\
...\\
...\\
+\\
\hline
r^f_t - \sum_{i=1}^{p'}\psi_i r^f_{t-i} &=  \sum_{i=0}^{q}\theta_i f_{t-i}\\
\end{split}
\end{equation}

The residual at time $t$ can be expressed as a sum of auto-regression over previous residual and the moving average of the features computed until $t$ as:

\begin{equation}
r^f_t = \underbrace{\sum_{i=1}^{p'}\psi_i r^f_{t-i}}_{auto-regression} + \underbrace{\sum_{i=0}^{q}\phi_i f_{t-i}}_{moving-average}
\end{equation}
or equivalently represented in the standard $ARMA(p',q)$ form as:
\begin{equation}
\begin{split}
(1 - \sum_{i=1}^{p'}\psi_i L^i)r^f_t &=  f_{t} + \sum_{i=1}^{q}\theta_i f_{t-i}\\
(1 - \sum_{i=1}^{p'}\psi_i L^i)r^f_t &=  (1 + \sum_{i=1}^{q}\theta_i L^i) f_{t}\\
\end{split}
\end{equation}
where, $L$ is the lag operator defined as $L^nf_t = f_{t-n}$. Now if the polynomial $(1-\sum _{i=1}^{p'}\alpha _{i}L^{i})$ has a unit root of multiplicity $d$, then

\begin{equation}
(1 - \sum_{i=1}^{p'}\psi_i L^i) = (1 - \sum_{i=1}^{p'-d}\phi_i L^i)(1-L)^d
\end{equation}

The residual series using feature $f$ can be expressed as an autoregressive integrated moving average process: 
\begin{equation}
ARIMA(p,d,q) = (1 - \sum_{i=1}^{p}\phi_i L^i) (1-L)^d r^f_t =  (1 + \sum_{i=1}^{q}\theta_i L^i) f_{t}\\
\end{equation}
where $p=p'-d$. While the distribution of $f_t$ terms is unknown, they can be generally assumed to be independent, identically distributed (i.i.d) variables. Intuitively, the residual is represented as an auto-regression of the previous $p$ residual terms and a moving average of the previous $q$ feature terms after $d$ non-seasonal differences (difference of lags that are required to make the series stationary) on the series. The variable in the series are:

\begin{itemize}
	\item $p$ - the number of autoregressive terms (the number of previous residual that are regressed up on)
	\item $q$ - the number of moving average terms (the number of previous features used in the moving average)
	\item $d$ - the number of non-seasonal differences (needed for stationarity)
	\item $\phi_i, i = \{1,2, ... ,p\}$ - the co-efficients for the $p$ autoregressive terms.
	\item $\theta_i, i = \{1,2, ... ,q\}$ - the co-efficients for the $q$ moving average terms.
\end{itemize}

We can apply a 1-D time series analysis if $f_i, t_t^f \in \Re$, This is true for features $f_i = \{b_2, b_4, e_1, e_3, e_4, k_1, k_2\}$. For $f_i = \{b_1, b_3, e_2\}$,  $f_i \rightarrow \Re^{MXN}$, and $r^f_t \rightarrow \Re$, where $MXN$ is the height and width of the image. In this case 1-D time series analysis can be applied if there exists a function ($F \rightarrow \Re$) that maps $f_i$ to a $\Re$, and satisfies the Cauchy's functional equation $F(a + b) = F(a) + F(b)$. In this case,
\begin{equation}
r^f_t = F(\alpha \bar{f}_t + \beta f_t) = \alpha F(\bar{f}_t) + \beta F(f_t) 
\end{equation}

\noindent The feature $b_1$, $b_3$, and $e_2$ are computed by applying a summation over all the pixel differences between the reference and test image that satisfies this condition. Hence, we can perform 1-D time series for these features as well.

\noindent\textbf{Estimating $\phi_i, \theta_j$}\\
Let’s assume that the order $(p, d, q)$ of the ARIMA series is known. An auto-regressive process without the moving average components is linear, and the parameters can be estimated using least squares regression. The moving average process introduces a non-linearity and requires non-linear estimation methods. The parameters of ARIMA models are estimated using the maximum likelihood estimation (MLE) method~\cite{box2015time}. However, a starting condition is required, which is estimated using a non-linear estimation method. Then MLE is conditioned on the starting point to solve the parameters of the ARIMA model~\cite{liu1992forecasting}. The MLE assumes that the error term has a normal distribution. While in our case, we do not know the distribution of the feature term. However, we estimate the parameters under the assumption that the features have a normal distribution. Features that are robust to sudden environment changes deviate less from the expected value and can be predicted well using the parameters estimated from the MLE. While on the other hand, features that have a complex distribution may not be robust to environment changes and are hard to predict and model. The prediction error from the model is used to ascertain the robustness of a feature.\\

\noindent\textbf{Model Selection ($p, q, d$})\\
The auto-correlation function ($ACF$) and the partial auto-correlation function ($PACF$) can be examined to approximately choose the order $(p, d, q)$. If $X_t$ is a random variable representing a stationary time series, then $ACF$ is defined as a function of lag ($h$) that computes the correlation between the random variables $X_t$ and $X_{t-h}$. The $PACF$ computes the correlation between $X_t$ and $X_{t-h}$ while filtering out the linear dependence of the random variables $\{X_{t-1}, X_{t-2}, ..., X_{t-h+1}\}$~\cite{box2015time}. This allows one to visually inspect the correlation plots to approximately choose the order of the model. 

Model selection tools such as Akaike's Information Criterion ($AIC$) is used to quantitatively select the order of the ARIMA models. The $AIC$ is a function of maximum likelihood and the number of parameters in the model. $AIC$ computes a score to quantify how well the model fits the data while penalizing the complexity of the model. This strikes a balance between selecting a model that fits the data well and is parsimonious~\cite{box2015time}. Given a stationary series, we estimate the parameters for various orders and compute their corresponding $AIC$. The model that minimizes the $AIC$ is chosen.

\vspace{-0.5cm}
\subsection{Analysis of Features}
Techniques for detecting tampering rely on the assumption that some features remain constant under normal operating conditions. Alternatively, we can argue that it is sufficient to model the change in the features of the image under normal operating conditions. To identify an image that has tampering, the techniques rely on the deviation of the feature from the model that defines the normal operating conditions. Considering a time series model, an ideal feature should produce residuals that are easy to model and are predictable while deviating sufficiently when there is tampering.
We quantify the robustness of a feature based on two characteristics:

\begin{itemize}
	\item Predictability of residuals under normal operation: Let $\{r^f_1, r^f_2, ... , r^f_n\}$ be a time series of residual computed for feature $f$ from time $\{0-t\}$. Let the time series be stationary and follow an $ARIMA(p,d,q)$ process, where $p,q,d$ are the optimal order that produces the minimum $AIC$. We define the robustness of the feature as the forecasting capability of the $ARIMA(p,d,q)$ process. The parameters of the $ARIMA(p,q,d)$ process are estimated by fitting the residual time series using a gradient descent optimization method~\cite{hyndman2007automatic}. Let $\{\hat{r}^f_{t}, \hat{r}^f_{t+1}, ... , \hat{r}^f_n\}$, be the predicted residual using the $ARIMA(p,d,q)$ process. The prediction error (deviation) is computed using the root means squared error ($RMSE$) between the residual and the predicted time series. To allow for comparison amongst the various features, a scale-independent RMSE ($sRMSE$) is computed as follows.
	
	\begin{equation}
	\label{eq:srmse}
	sRMSE = \frac{\sqrt{\frac{1}{n-t}\sum_{k=t }^{n} (r^f_k - \hat{r}^f_k)^2}}{(max_{k}(r^f_k) - min_k(r^f_k))}
	\end{equation}
	
	The auto-regressive process tends to converge to the arithmetic mean of the series over long-term predictions. We perform an in-sample prediction. To predict the residual at time $k$, we consider the previous residual until $k-1$ rather than considering the out of sample predictions. 
	\item Deviation of residuals under tampering: Let $\{I_1, I_2, ..., I_t\}$ be images captured from a surveillance camera that is functioning normally at time instants $t = \{1,2, ..., t\}$. Hypothetically, let $\{I^\tau_1, I^\tau_2, ..., I^\tau_t\}$ be the images from the same surveillance camera when tampered, at time instants $t = \{1,2, ..., t\}$. (Note that only one of these two sets of images can occur in the real world. We perform analysis through synthesizing tampering into the normal images allowing the two sets to be available). The deviation of the residual under tampering from the residuals under normal operation can be quantified using $sRMSE$ (Eq.~\ref{eq:srmse}) as well.

\end{itemize}
A robust feature would produce a small $sRMSE$ computed over predictability and a large value for the $sRMSE$ computed over the deviations.
\vspace{-0.3cm}
\subsection{Performance of Features}
Given an annotated dataset and the corresponding residuals for a feature $f$, we compute the receiver operating characteristic (ROC) curve, and the area under the ROC curve (AUC ROC) to ascertain the classification capability using a certain feature. We pick an optimal threshold, defined as the point on the ROC where the distance between the true positive rate (TPR) and false-positive rate (FPR) is largest, and analyze the confusion matrix, the accuracy, and the F1 score to compare the performance using different features. 
\vspace{-0.3cm}
\subsection{Experiments}

\noindent\textbf{Dataset}\\
We use videos from the University of Houston Camera Tampering Detection (UHCTD)~\cite{mantini2019uhctd} dataset to conduct experiments. UHCTD is representative of video captured from a surveillance camera and contains synthetic tampering for evaluation. The dataset consists of a video from a surveillance camera captured over a period of $24$ hours at $3$ fps. The images are originally captured at $1080p$ resolution.  We synthesize the data required for evaluation, using image processing techniques similar to the method described in ~\cite{mantini2019uhctd}. Four classes of data are synthesized (normal, covered, defocussed, and moved). 



\begin{figure}
	\centering
	
	\begin{subfigure}{}
		\includegraphics[width=0.1\textwidth]{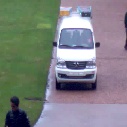}
	\end{subfigure}
	\begin{subfigure}{}
		\includegraphics[width=0.1\textwidth]{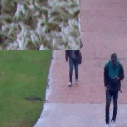}
	\end{subfigure}
	\begin{subfigure}{}
		\includegraphics[width=0.1\textwidth]{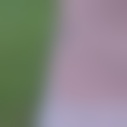}
	\end{subfigure}
	\begin{subfigure}{}
		\includegraphics[width=0.1\textwidth]{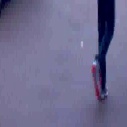}
	\end{subfigure}
	
	\begin{subfigure}{}
		
	\end{subfigure}
	
	\begin{subfigure}{}
		\includegraphics[width=0.1\textwidth]{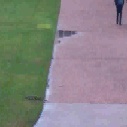}
	\end{subfigure}
	\begin{subfigure}{}
		\includegraphics[width=0.1\textwidth]{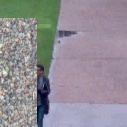}
	\end{subfigure}
	\begin{subfigure}{}
		\includegraphics[width=0.1\textwidth]{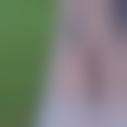}
	\end{subfigure}
	\begin{subfigure}{}
		\includegraphics[width=0.1\textwidth]{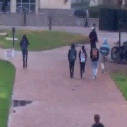}
	\end{subfigure}
	
	\begin{subfigure}{}
		
	\end{subfigure}
	
	\begin{subfigure}{}
		\includegraphics[width=0.1\textwidth]{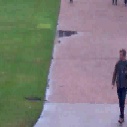}
	\end{subfigure}
	\begin{subfigure}{}
		\includegraphics[width=0.1\textwidth]{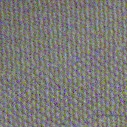}
	\end{subfigure}
	\begin{subfigure}{}
		\includegraphics[width=0.1\textwidth]{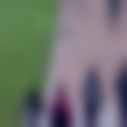}
	\end{subfigure}
	\begin{subfigure}{}
		\includegraphics[width=0.1\textwidth]{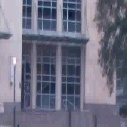}
	\end{subfigure}
	
	\begin{subfigure}{}
		(a) left, (b) mid-left, (c) mid-right, (d) right
	\end{subfigure}
	\caption{Synthetic Data. a) Original, b) Covered, c) Defocussed, and d) Moved images}\label{fig:synthetic_data}
	
\end{figure}

Two datasets are synthesized from the same video for analysis; the first one containing normal images, and the second one containing a combination of four classes of tampered images (normal, covered, defocussed, and moved).
\begin{itemize}
	\item Normal Dataset: This dataset consists of over $250K$ normal images.
	\item Tampered Dataset: This dataset consists of four classes of images: normal, covered, defocussed, and moved. Similar to ~\cite{mantini2019uhctd} we induced tampering every ten minutes over a period of 24 hours. Each tampering last between five to ten minutes. We vary two parameters to induce tampering: extent, and rate~\cite{mantini2019uhctd}. The dataset consists of over $250K$ images of which a quarter ($65K$) have tampering; the tampers are distributed uniformly (approximately $21K$ of each type).
\end{itemize}
Figure~\ref{fig:synthetic_data} shows examples of normal and tampered images in the dataset.

\noindent\textbf{Stationarity analysis}\\
$ARIMA$ process assumes that the time series is stationary. A process is stationary if the unconditional joint probability distribution of any sequence of random variables in the series is constant. We relax this condition and test the residual time series for 2\textsuperscript{nd} order stationarity. A series is stationary to the order 2 if the series displays a constant mean, variance, and a finite time-invariant auto-covariance. For a non-stationary process, we apply an appropriate transformation on the series to induce stationarity.

Test for unit root: If the characteristic equation of a time series has a unit root, then it can be shown that the series has a variance that is dependent on time, and hence is non-stationary. Furthermore, a series with a root greater than one is explosive and cannot be modeled. Stationary processes have roots less than 1. We test the unit root hypothesis using the Augmented Dickey-Fuller (ADF) test~\cite{hamilton1994time}. The intuition for the test is that, if a series is stationary, the change in $y_t$ ($\triangle~y_t$) will not depend on the lagged level of the series ($y_{t-1}$). ADF performs a t-test with the null hypothesis that the series has a unit root and the alternative that the series does not have a unit root. Accepting the alternative implies that the series is either stationary or trend stationary. As the variables of the test could be non-stationary, the t-statistics are not compared with the critical values for a standard t-distribution, but rather with the critical values calculated by Dickey and Fuller~\cite{dickey1979distribution}. If the t-value is less than the critical value, the null hypothesis is rejected. 

Test for stationarity: The absence of a unit root does not imply stationarity. It is possible for a series to be non-stationary, have no unit root, and yet be trend-stationary. We employ Kwiatkowski Phillips Schmidt Shin (KPSS) test~\cite{kwiatkowski1992testing} to determine if the series is stationary around a mean or linear trend. KPSS is a regression-based test with the null hypothesis that the series is stationary. Contrary to most test, the alternative hypothesis is that the series has a unit root. We accept the null hypothesis that the series is stationary or trend-stationary if the test statistic is less than the chosen critical value.

\begin{table}[]
	\center
	\begin{tabular}{|c||c|c||l|l|}
		\hline
		Feature & \multicolumn{2}{c||}{ADF}  & \multicolumn{2}{c|}{KPSS} \\ \cline{2-5}
		& ($a$) &  $H_0$ & ($k$) & $H_0$ \\

		\hline
		$b_1$& \textbf{-41.912} & Reject & 0.613 & Reject \\
		$b_2$& -15.195 & Reject & 65.553 & Reject \\
		$b_3$& -39.338 & Reject & \textbf{0.161} & Accept \\
		$b_4$& -17.255 & Reject & 19.908 & Reject \\
		$e_1$& -12.594 & Reject & 126.037 & Reject \\
		$e_2$& -14.314 & Reject & 117.416 & Reject \\
		$e_3$& -12.594 & Reject & 126.037 & Reject \\
		$e_4$& -7.293 & Reject & 131.937 & Reject \\
		$k_1$& -27.044 & Reject & 2.152 & Reject \\
		$k_2$& -15.363 & Reject & 93.074 & Reject\\
		\hline
	\end{tabular}

	\begin{tabular}{|c|}

	\end{tabular}

	\begin{tabular}{|c||c|c||l|l|}
		\hline
		Feature & \multicolumn{2}{c||}{ADF}  & \multicolumn{2}{c|}{KPSS} \\ \cline{2-5}
		& ($a$) &  $H_0$ & ($k$) & $H_0$ \\
		
		\hline
		$b_1$& -78.767 & Reject & $1.9X10^{-4}$ & Accept \\
		$b_2$& -77.913 & Reject & $6.2X10^{-4}$ & Accept \\
		$b_3$& -64.960 & Reject & $2.9X10^{-4}$ & Accept \\
		$b_4$& -81.960 & Reject & $12X10^{-4}$ & Accept \\
		$e_1$& -78.465 & Reject & $9.4X10^{-4}$ & Accept \\
		$e_2$& -86.074 & Reject & $26X10^{-4}$ & Accept \\
		$e_3$& -78.465 & Reject & $9.4X10^{-4}$ & Accept \\
		$e_4$& -77.624 & Reject & $60X10^{-4}$ & Accept \\
		$k_1$& -84.800 & Reject & $1.9X10^{-4}$ & Accept \\
		$k_2$& -79.018 & Reject & $6.7X10^{-4}$ & Accept\\
		\hline
	\end{tabular}
	\caption{(Top: Before transformation, Bottom: After transformation) ADF test for unit root: $H_0$ - has unit root, if ADF statistic $a < -2.861$ ($5\%$ critical value), then Reject $H_0$; and KPSS test for stationarity: $H_0$: is stationary, if KPSS statistic $k < 0.463$ ($5\%$ critical value), then Accept $H_0$. \label{tab:adfkpss}}
\end{table}

Analysis: The test results and statistics are shown in Table~\ref{tab:adfkpss} (Top).
\begin{itemize}
	\item The null hypothesis for the ADF test that the series has a unit root is rejected for all the features. (The alternative is accepted, that the series does not have a unit root)
	\item The smaller the value of the ADF test statistic, the stronger is the rejection of the null hypothesis. The test shows that feature $b_1$, and $b_3$ are unlikely to have a unit root. 
	\item The null hypothesis for the KPSS test that the series is stationary or trend-stationary is rejected for all the features except for $b_3$. 
	\item Feature $b_3$ is computed as a time delayed difference between the backgrounds ($b_{t} - b_{t-n}$), which is a difference of lag $n$ on the features. Differencing a series often results in a stationary series.
\end{itemize}

\noindent Pre-processing for stationarity: We induce stationarity for further analysis by applying a log transformation and lag differencing. If $\{r^f_1, r^f_2, ... , r^f_n\}$ is a residual feature, then we apply trasformation as a first order lag difference on the series $\{log(r^f_1+min_f), log(r^f_2+min_f), ... , log(r^f_n+min_f)\}$, where $min_f = |min_{i}\{f_i\}|$ if the series has negative values, otherwise $min_f=0$. The ADF and KPSS test results for the transformed series are shown in Table~\ref{tab:adfkpss} (Bottom) and indicate that the transformed series are stationary. 

\noindent\textbf{Robustness of feature:}\\
We quantify the predictability of the residuals under normal operating conditions and the deviation of the residual under tampering on the synthetic dataset.

\begin{figure}
	\centering
	
	\begin{subfigure}{}
		\includegraphics[width=0.1\textwidth]{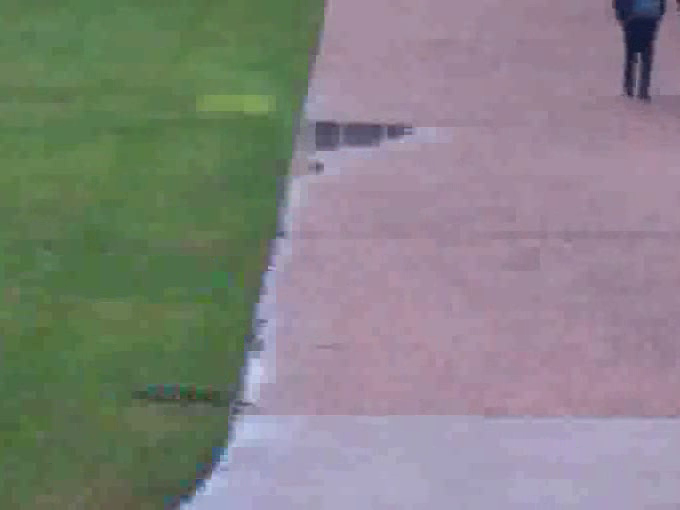}		
	\end{subfigure}
	\begin{subfigure}{}
		\includegraphics[width=0.1\textwidth]{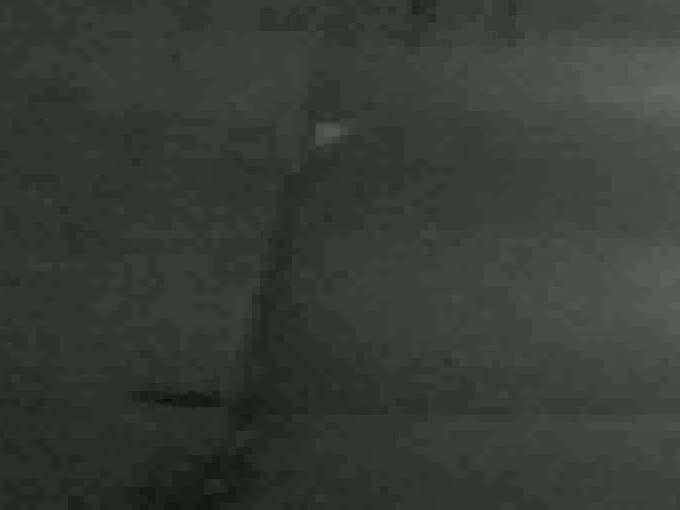}
	\end{subfigure}
	\begin{subfigure}{}
		\includegraphics[width=0.1\textwidth]{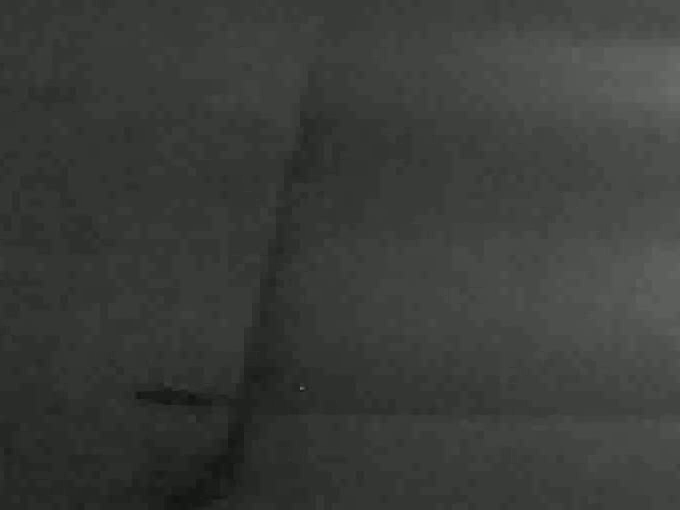}
	\end{subfigure}
	\begin{subfigure}{}
		\includegraphics[width=0.1\textwidth]{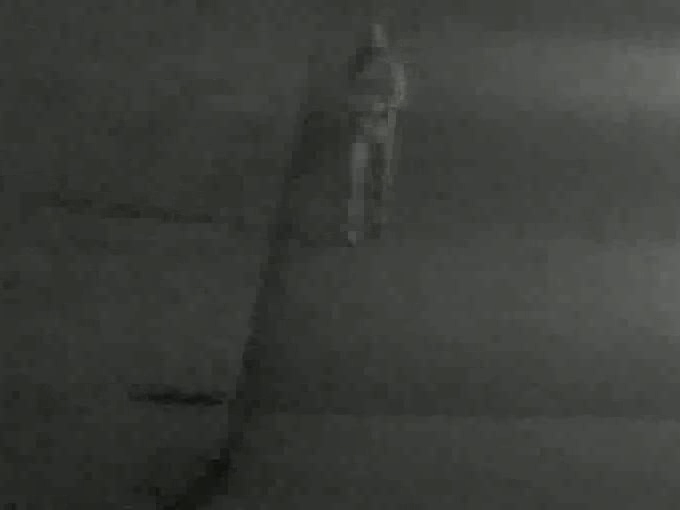}
	\end{subfigure}
	
	\begin{subfigure}{}
		
	\end{subfigure}
	
	\begin{subfigure}{}
		\includegraphics[width=0.1\textwidth]{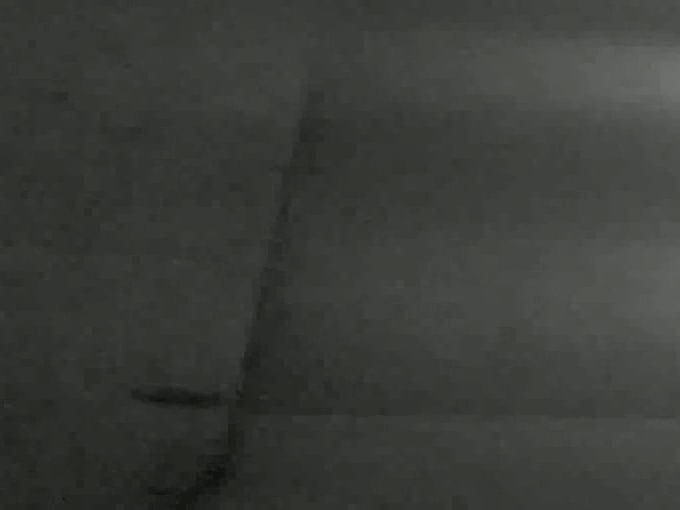}
	\end{subfigure}
	\begin{subfigure}{}
		\includegraphics[width=0.1\textwidth]{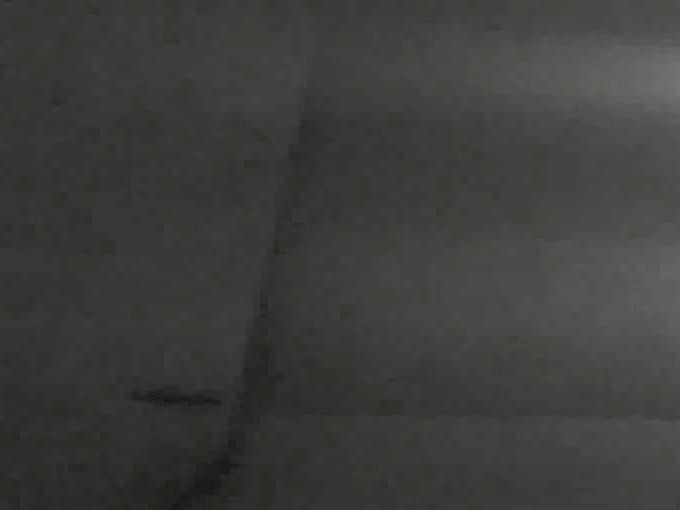}
	\end{subfigure}
	\begin{subfigure}{}
		\includegraphics[width=0.1\textwidth]{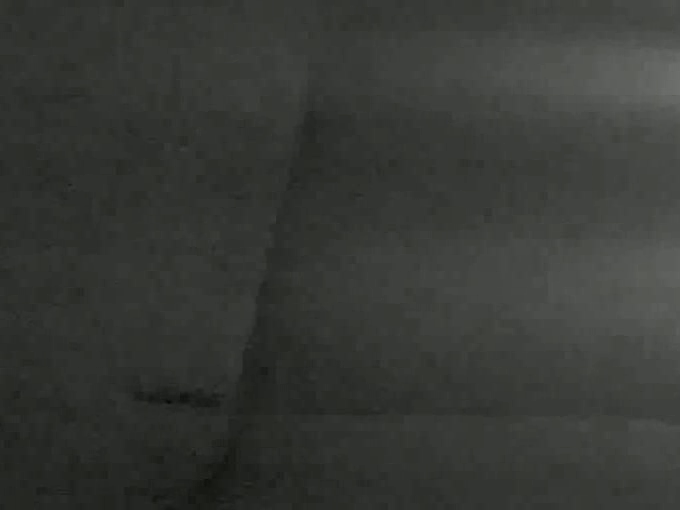}
	\end{subfigure}
	\begin{subfigure}{}
		\includegraphics[width=0.1\textwidth]{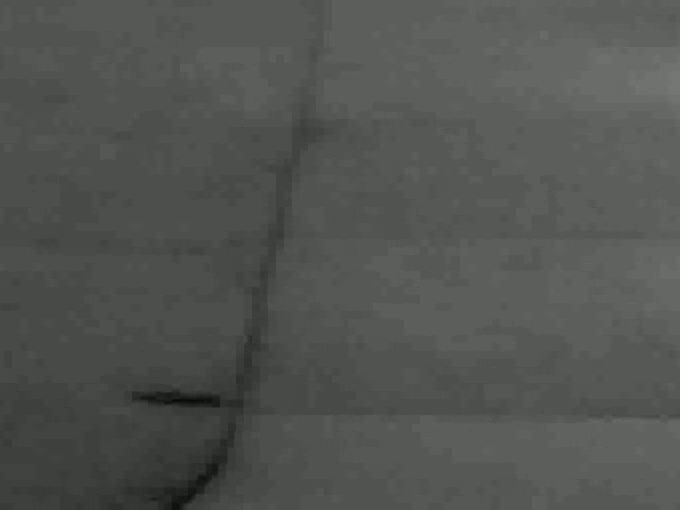}
	\end{subfigure}
	
	\begin{subfigure}{}
		
	\end{subfigure}
	
	\begin{subfigure}{}
		\includegraphics[width=0.1\textwidth]{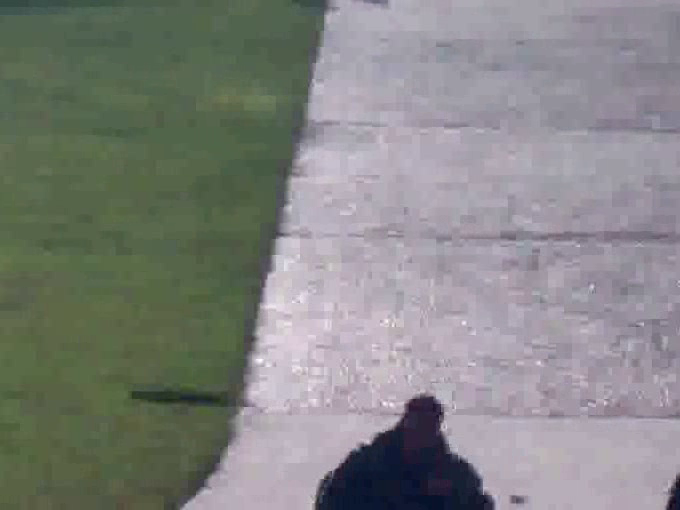}
	\end{subfigure}
	\begin{subfigure}{}
		\includegraphics[width=0.1\textwidth]{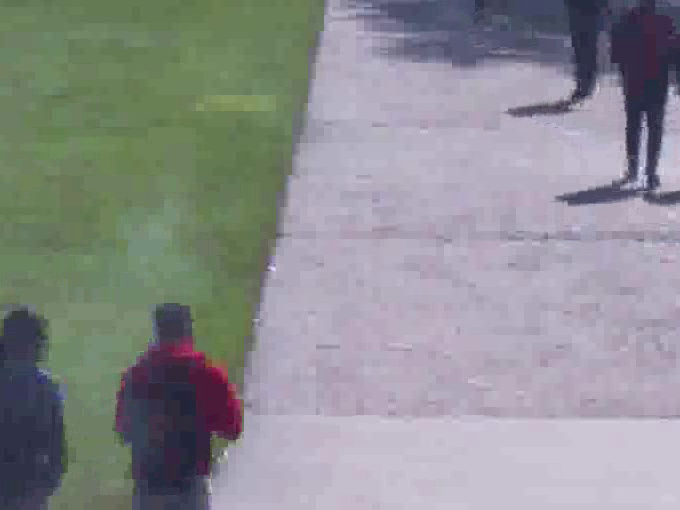}
	\end{subfigure}
	\begin{subfigure}{}
		\includegraphics[width=0.1\textwidth]{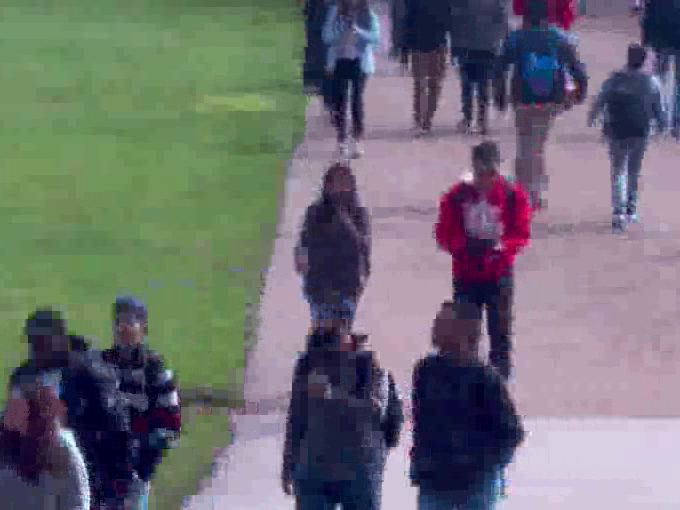}
	\end{subfigure}
	\begin{subfigure}{}
		\includegraphics[width=0.1\textwidth]{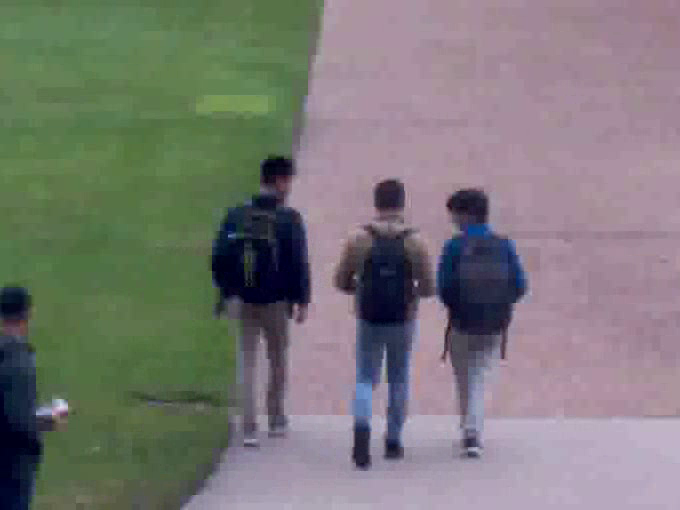}
	\end{subfigure}
	
	\begin{subfigure}{}
		(Top Row) Hour 1, 3, 5, 7; \\(Middle Row) Hour 9, 11, 13, 15; \\(Bottom Row) Hour 17, 19, 21, 23
	\end{subfigure}
	\caption{Normal Dataset. Scene variability over time}\label{fig:normal_data}
	
\end{figure}

ARIMA(p,d,q) order analysis: We use the Normal dataset to perform this analysis. The normal dataset consists of a video from a surveillance camera under normal operating conditions, over a period of 24 hours. There is large variability in the global illumination and scene over this period as shown in Figure~\ref{fig:normal_data}. We assume that the $ARIMA$ process is different for each period of time. We build and estimate the parameters of $ARIMA$ model for each hour of the video.

\begin{table*}[th]
	\tiny
	\centering
	\begin{tabular}{|p{.75cm}|p{.20cm}p{.20cm}p{.20cm}p{.20cm}p{.20cm}p{.20cm}p{.20cm}p{.20cm}p{.20cm}p{.20cm}p{.20cm}p{.20cm}p{.20cm}p{.20cm}p{.20cm}p{.20cm}p{.20cm}p{.20cm}p{.20cm}p{.20cm}p{.20cm}p{.20cm}p{.20cm}p{.30cm}|}
		
		\hline
		feature/hour& 1 & 2 & 3 & 4 & 5 & 6 & 7 & 8 & 9 & 10 & 11 & 12 & 13 & 14 & 15 & 16 & 17 & 18 & 19 & 20 & 21 & 22 & 23 & 24 \\	
		\hline
		$b_1$&(4,1,4)&(1,1,1)&(2,1,5)&(6,1,3)&(5,1,3)&(5,1,5)&(0,1,0)&(3,1,3)&(2,1,5)&(4,1,6)&(5,1,2)&(6,1,4)&(4,1,4)&(2,1,4)&(5,1,5)&(9,1,1)&(3,1,0)&(4,1,2)&(5,1,4)&(4,1,4)&(4,1,3)&(5,1,3)&(4,1,4)&(3,1,1)\\
		$b_2$&(3,1,1)&(3,1,2)&(4,1,4)&(2,1,1)&(0,1,0)&(4,1,6)&(6,1,3)&(2,1,2)&(4,1,5)&(4,1,6)&(3,1,2)&(2,1,5)&(1,1,5)&(4,1,2)&(3,1,4)&(5,1,5)&(5,1,2)&(3,1,3)&(7,1,3)&(1,1,0)&(5,1,4)&(4,1,3)&(3,1,2)&(2,1,0)\\
		$b_3$&(6,1,4)&(3,1,7)&(6,1,4)&(5,1,5)&(5,1,3)&(6,1,4)&(3,1,2)&(3,1,3)&(5,1,5)&(3,1,3)&(4,1,6)&(4,1,6)&(5,1,5)&(3,1,3)&(4,1,3)&(6,1,4)&(6,1,4)&(4,1,3)&(6,1,4)&(5,1,5)&(5,1,5)&(6,1,4)&(6,1,4)&(6,1,4)\\
		$b_4$&(3,1,1)&(4,1,6)&(4,1,6)&(5,1,5)&(1,1,4)&(6,1,3)&(2,1,3)&(3,1,6)&(2,1,8)&(1,1,3)&(1,1,2)&(1,1,2)&(4,1,6)&(2,1,4)&(3,1,4)&(1,1,3)&(2,1,4)&(2,1,3)&(1,1,2)&(1,1,2)&(1,1,4)&(5,1,5)&(1,1,3)&(1,1,2)\\
		$e_1$&(3,1,1)&(5,1,5)&(3,1,6)&(5,1,5)&(0,1,1)&(5,1,2)&(5,1,5)&(5,1,5)&(1,1,5)&(0,1,1)&(6,1,4)&(5,1,5)&(4,1,6)&(0,1,2)&(5,1,5)&(3,1,3)&(4,1,5)&(3,1,6)&(4,1,5)&(3,1,5)&(3,1,3)&(3,1,4)&(2,1,3)&(3,1,3)\\
		$e_2$&(4,1,3)&(4,1,3)&(5,1,5)&(3,1,3)&(6,1,3)&(1,1,1)&(2,1,2)&(6,1,4)&(3,1,5)&(3,1,4)&(3,1,4)&(3,1,4)&(4,1,6)&(4,1,5)&(4,1,6)&(4,1,5)&(5,1,2)&(3,1,4)&(3,1,6)&(3,1,3)&(1,1,1)&(7,1,3)&(2,1,4)&(3,1,4)\\
		$e_3$&(3,1,1)&(5,1,5)&(3,1,6)&(5,1,5)&(0,1,1)&(4,1,4)&(5,1,5)&(5,1,3)&(1,1,5)&(0,1,1)&(6,1,4)&(5,1,5)&(4,1,6)&(0,1,2)&(5,1,5)&(3,1,3)&(4,1,5)&(3,1,6)&(4,1,5)&(3,1,5)&(3,1,3)&(3,1,4)&(2,1,3)&(3,1,3)\\
		$e_4$&(5,1,5)&(4,1,4)&(6,1,3)&(3,1,5)&(7,1,3)&(5,1,5)&(6,1,4)&(6,1,4)&(3,1,6)&(3,1,6)&(4,1,4)&(3,1,6)&(6,1,4)&(4,1,6)&(5,1,5)&(5,1,5)&(4,1,6)&(6,1,4)&(2,1,2)&(0,1,0)&(6,1,4)&(5,1,2)&(2,1,4)&(7,1,3)\\
		$k_1$&(0,1,2)&(3,1,5)&(4,1,3)&(3,1,3)&(1,1,2)&(5,1,5)&(2,1,4)&(4,1,4)&(1,1,1)&(4,1,6)&(1,1,1)&(5,1,5)&(1,1,4)&(4,1,5)&(4,1,3)&(3,1,4)&(1,1,0)&(3,1,4)&(5,1,4)&(1,1,2)&(0,1,1)&(4,1,5)&(5,1,5)&(5,1,5)\\
		$k_2$&(2,1,3)&(0,1,1)&(3,1,3)&(1,1,1)&(6,1,4)&(4,1,3)&(5,1,5)&(4,1,1)&(0,1,1)&(3,1,5)&(4,1,3)&(3,1,3)&(5,1,1)&(4,1,4)&(3,1,3)&(3,1,4)&(2,1,4)&(3,1,5)&(3,1,2)&(3,1,2)&(3,1,4)&(3,1,3)&(3,1,4)&(1,1,1)\\
		\hline
	\end{tabular}
	\caption{Estimate Order $(p,d,q)$ of the $ARIMA$ process for each features by minimizing the $AIC$ \label{tab:order}}
	
\end{table*}

For completeness of the analysis, the order of the $ARIMA$ process for each feature is shown in Table.~\ref{tab:order}. The order $d=1$, as we apply a difference of lag 1 on the features to make them stationary. The average autoregressive term and moving average terms that minimize the $AIC$ are 3 and 4, respectively. We perform students’ t-test to ascertain if there is a significant difference in the average number of auto-regressive and the moving average terms to model the $ARIMA$ process.
\begin{itemize}
	\item  The number of optimal parameters required for $b_3$ and $e_4$ were the highest (mean AR terms: 4.79, mean MA terms: 4.16)
	\item Features $b_4$ and $k_1$ requires the least number of AR parameters, 2.37, and 2.87, respectively.
	\item Features $b_2$ and $k_2$ require the least number of MA parameters, $2.19$ each.
	\item For this dataset, the complexity of modeling, and predicting the residuals is maximum for $b_3$ and $e_4$, and minimum for $k_2$. 
	\item Tests suggest no significant difference in the average optimal order for the background features, and edge features, however the same is lower for the keypoint based features.
	\item In general, to make a decision based on choice, one would have to consider the complexity of the feature computation as well. Extracting keypoints is more complex than detecting background or edges. 
\end{itemize}

%
%

Predictability of residuals under normal operating conditions: To quantify the forecast error for each feature, we fit the residuals for each hour of the video based on the order estimated in the previous step. Then an insample prediction for the last 1000 time instances is performed. The mean $sRMSE$ over the 24 hour period is shown in Table~\ref{tab:pred_srmse} (Top).

\begin{table*}[th]	
	
	\centering
	\begin{tabular}{|c|c|c|c|c|c|c|c|c|c|c|}		
		\hline
		feature& $b_1$ & $b_2$ & $b_3$ & $b_4$ & $e_1$ & $e_2$ & $e_3$ & $e_4$ & $k_1$ & $k_2$\\	
		\hline
		$b_1$&--& 0.072& 0.005& 5xe-6& 0.861& 0.001& 0.863& 0.129& 0.179& 0.012\\
		$b_2$& 0.072& --& 0.010& 0.346& 0.084& 0.190& 0.084& 0.319& 0.953& 0.811\\
		$b_3$&0.005& 0.010& --& 6xe-9& 0.006& 9xe-5& 0.006& 0.001& 0.053& 1xe-4\\
		$b_4$&5xe-6& 0.346& 6xe-9& --& 1xe-5& 0.465& 1xe-5& 0.003& 0.407& 0.085\\
		$e_1$&0.861& 0.084& 0.006& 1xe-5& --& 0.001& 0.998& 0.176& 0.198& 0.017\\
		$e_2$&0.001& 0.190& 9xe-5& 0.465& 0.001& --& 0.001& 0.012& 0.230& 0.073\\
		$e_3$&0.863& 0.084& 0.006& 1xe-5& 0.998& 0.001& --& 0.175& 0.197& 0.017\\
		$e_4$&0.129& 0.319& 0.001& 0.003& 0.176& 0.012& 0.175& --& 0.467& 0.256\\
		$k_1$&0.179& 0.953& 0.053& 0.407& 0.198& 0.230& 0.197& 0.467& --& 0.900\\
		$k_2$&0.012& 0.811& 0.000& 0.085& 0.017& 0.073& 0.017& 0.256& 0.900& --\\
		\hline
	\end{tabular}
	\caption{p-value of t-tests between prediction error of different features over 24 hours, $p \le 0.05$ implies significant difference in mean \label{tab:predictability}}
	
\end{table*}

\begin{itemize}
	\item Feature $b_3$ produces the least error on forecasting the residuals. Results from t-tests (Tab~\ref{tab:predictability}) suggest a difference in the mean of the $sRMSE$ (computed over the 24 hour period) for $b_3$ from all the other features except $k_1$. For this dataset, we conclude that the reference feature $b_3$ can be modeled and predicted with a higher accuracy compared to $\{b_1, b_2, b_4, e_1, e_2, e_3, e_4, k_2\}$.
	\item Feature $e_2$ produces the highest error in forecasting the residuals. Results from t-tests suggest the difference in mean is only significant for $\{b_1, b_3, e_1, e_3, e_4\}$.
	\item Results from t-test do not show any significance in the mean of the forecasting error for background, edge, and key-point features when compared as a whole. 
\end{itemize}


\noindent Deviation of residuals under tampering: To quantify the deviation under tampering, we accumulate the tampered images from the Tampered dataset into three groups: covered, defocussed, and moved. For each frame that has tampering, we accumulate the corresponding frames from the Normal dataset. The sRMSE is computed between the residual computed over the tampered frame and the normal frame. Table~\ref{tab:pred_srmse} (Bottom) shows the sRMSE score for each group of tampering.


\begin{table*}[]

	\center
	\begin{tabular}{|c||l|}
		\hline
		Feature & sRMSE \\
		\hline
		$b_1$& 0.390\\
		$b_2$& 1.044\\
		$b_3$& \textbf{0.122}\\
		$b_4$& 1.414\\
		$e_1$& 0.413\\
		$e_2$& \textbf{1.718}\\
		$e_3$& 0.413\\
		$e_4$& 0.664\\
		$k_1$& 1.011\\
		$k_2$& 0.949\\
		\hline
	\end{tabular}
	\begin{tabular}{|c||l|l|l|}
	\hline	
	Feature/sRMSE & Covered & Defocussed & Moved\\
	\hline
	$b_1$& 0.3483 & 0.1420 & 0.0864 \\
	$b_2$& 0.2665 & 0.1190 & 0.1802 \\
	$b_3$& 0.7154 & 0.1669 & \textbf{0.3540} \\
	$b_4$& 0.1631 & 0.1153 & 0.1918 \\
	$e_1$& 0.4178 & 0.1880 & 0.1696 \\
	$e_2$& 1.0810 & \textbf{0.2846} & 0.3275 \\
	$e_3$& 0.4178 & 0.1880 & 0.1696 \\
	$e_4$& 0.2447 & 0.1213 & 0.1527 \\
	$k_1$& \textbf{4.4555} & 0.0490 & 0.0744 \\
	$k_2$& 0.1237 & 0.0455 & 0.0447 \\
	\hline
	\end{tabular}
	\caption{(Top) Prediction under normal operating conditions: sRMSE. (Bottom) Deviation under tampering: sRMSE \label{tab:pred_srmse}}
\end{table*}

\begin{itemize}
	\item Feature $k_1$ produces the largest deviation for covered tampering, $e_2$ produces the largest deviation for defocussed, and $b_3$ produces the largest deviation for moved tampering. 
	\item The deviation of the residual for $k_1$ due to covered tampering is much larger compared to the deviation in any other feature from any tampering. The residual for $k_1$ is computed as a difference in the number of key-points. This large deviation could be attributed to the fact that covered tampering is synthesized by replacing a region of the image with random texture, which could generate a large number of key-points. It is uncertain if the large deviation would persist if the region was to be replaced by uniform color pixels.
	\item Excluding $k_1$ from the argument, feature $e_2$ produces a significant deviation for covered, defocussed, and moved tampering. 
\end{itemize}

\begin{figure*}
	\centering
	
	\begin{subfigure}{}
		\includegraphics[width=0.31\textwidth]{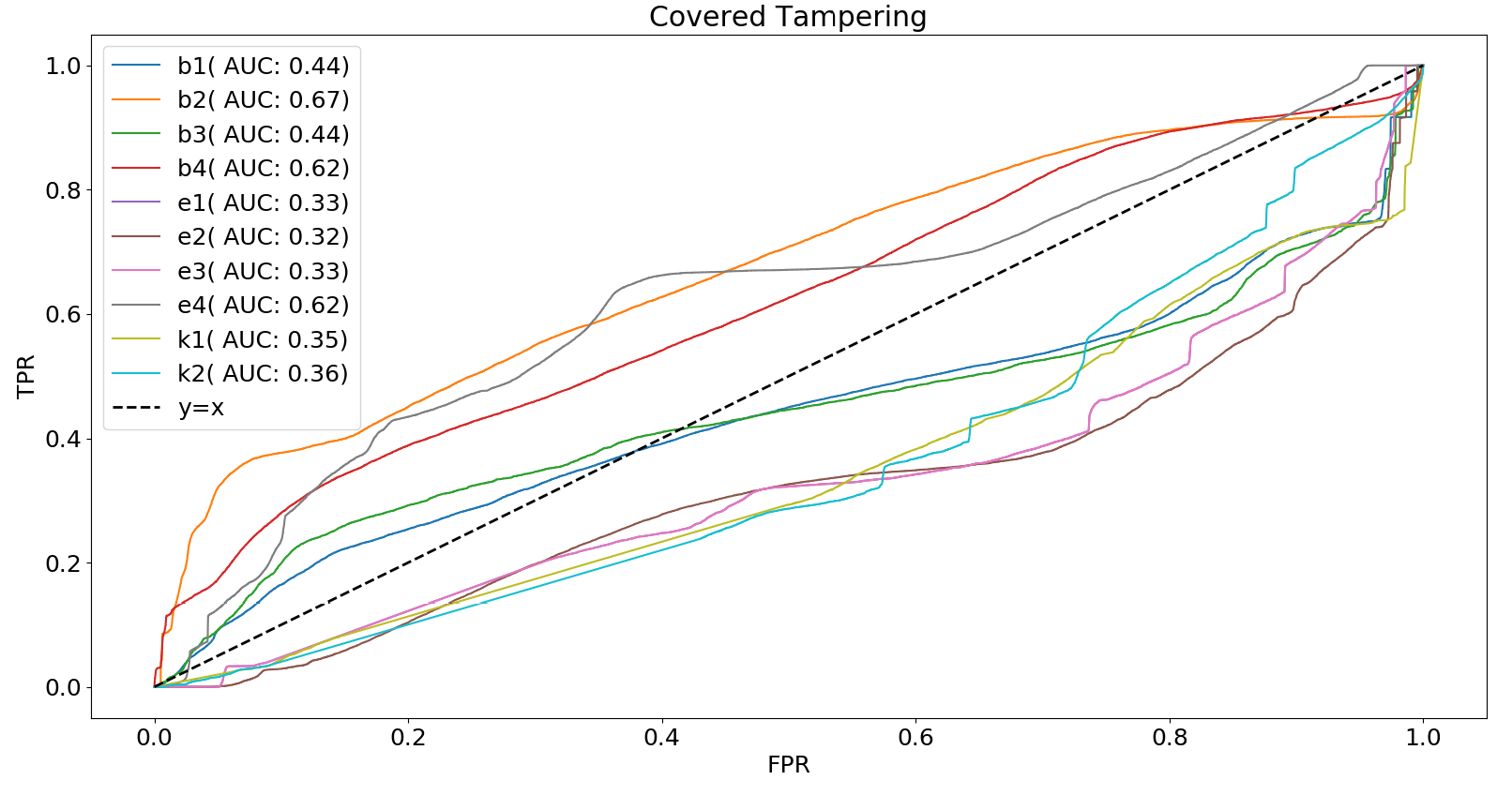}		
	\end{subfigure}
	\begin{subfigure}{}
		\includegraphics[width=0.31\textwidth]{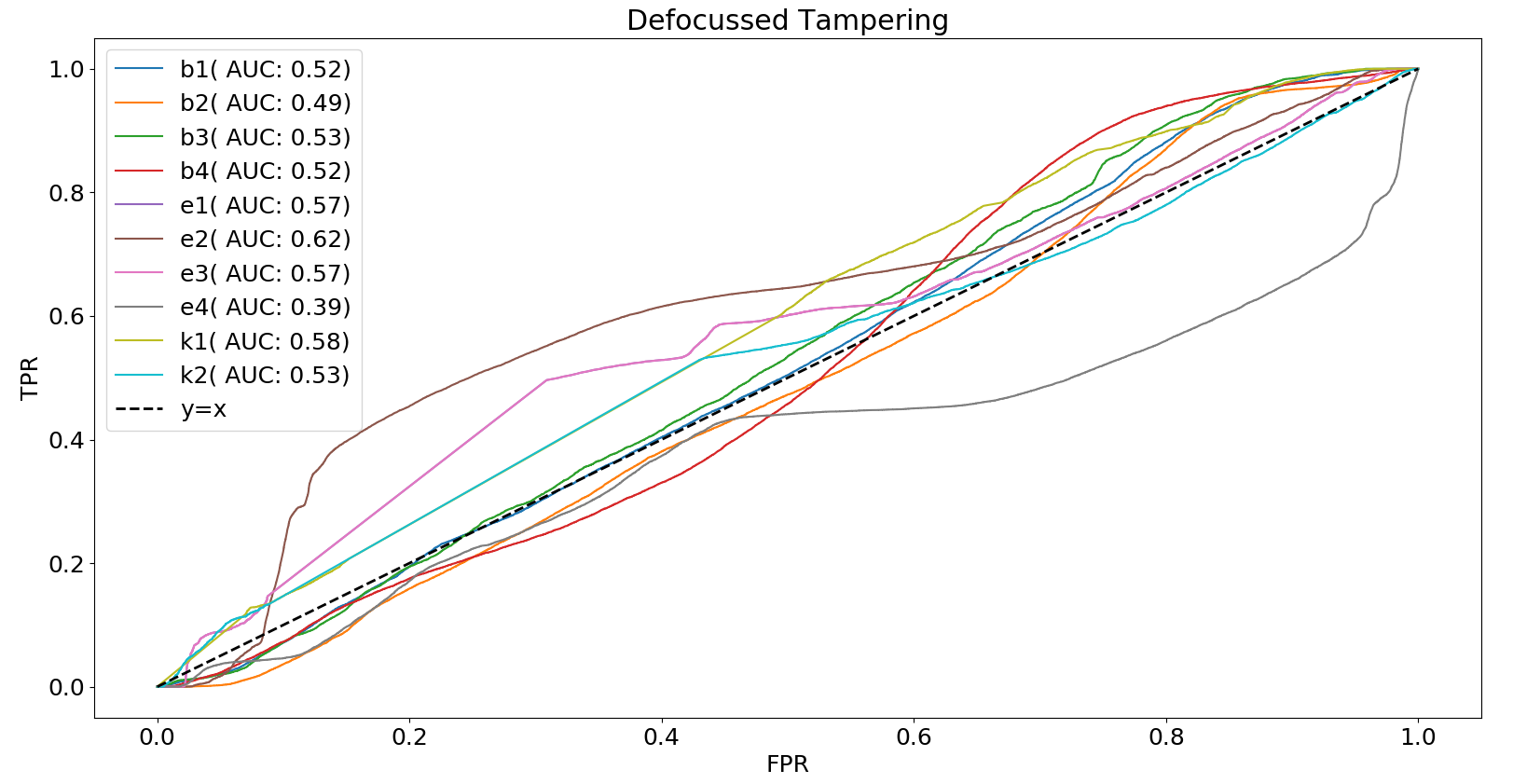}
	\end{subfigure}
	\begin{subfigure}{}
		\includegraphics[width=0.31\textwidth]{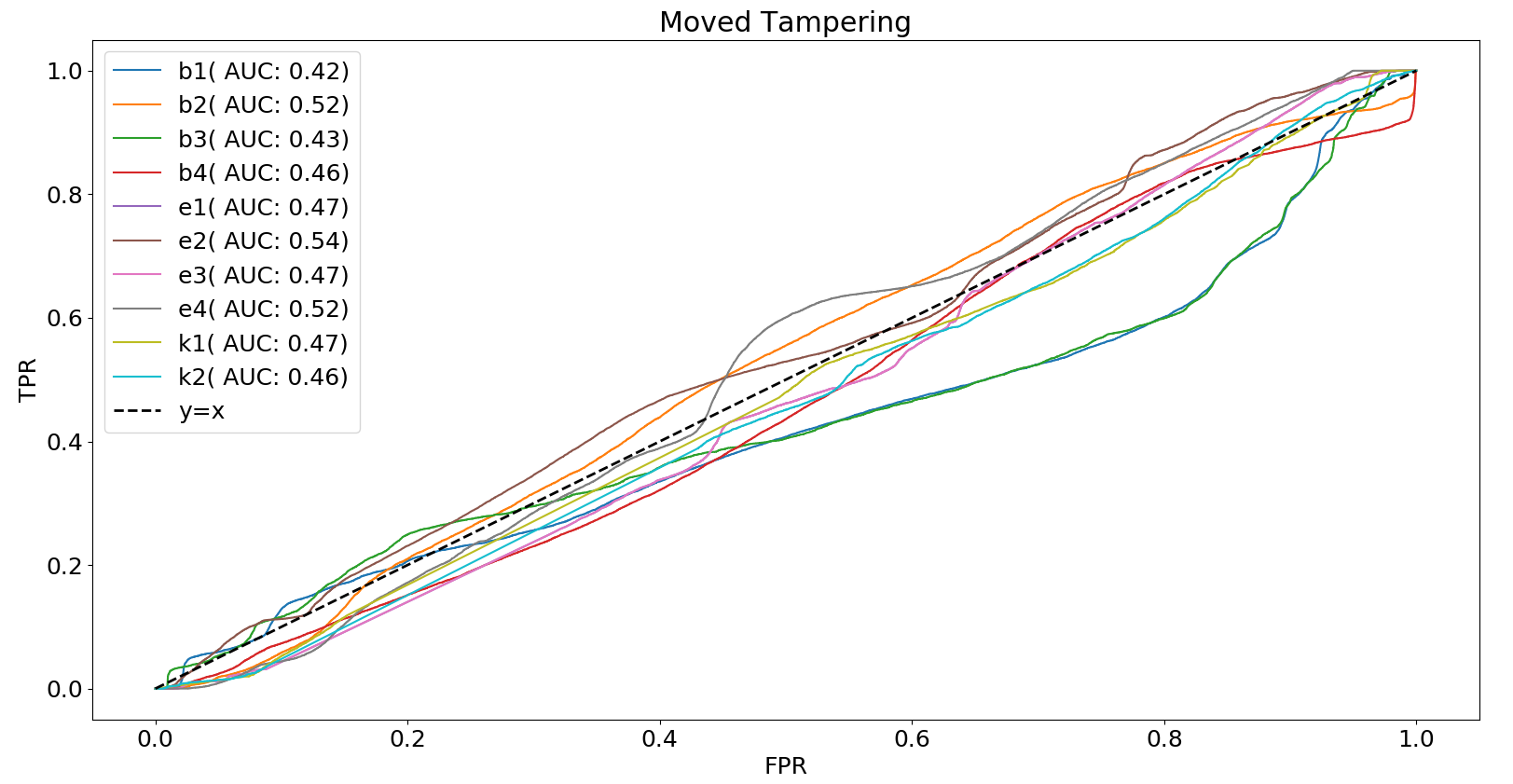}
	\end{subfigure}
	\begin{subfigure}{}

	\end{subfigure}
	\caption{(Left) Covered Tampering; (Center) Defocussed Tampering; (Right) Moved Tampering; 
		ROC and AUC compute over the tampering dataset}\label{fig:roc}
	
\end{figure*}

\noindent\textbf{Performance of feature:}\\
We analyze the detection capability of various features by thresholding the residuals. We perform this analysis using the tampering dataset. For each feature, the residuals are computed on the tampering dataset. The ground truth annotation for the tampering in the video is annotated during synthesis. The performance is quantified for each feature on three groups of data, where each group contains only one type of tampering. The Receiver Operating Characteristic curves along with the corresponding area under the curve are shown in Fig.~\ref{fig:roc}

 \begin{itemize}
	\item The ROC curves suggest that feature $b_2$, $b_4$, and $e_4$ are able to detect covered tampering better than other features. 
	\item Feature $b_2$ (AUC=0.67) captures the difference in entropy between the background and the current image and outperforms other features in detecting covered tampering. 
	\item Feature $b_4$ (AUC=0.62) captures the difference in the concentration around the maximum values in the histogram and $e_4$ (AUC=0.62) captures the intersection of edges between reference and test image and performs covered tapering detection equally well.
	\item Feature $e_2$ (AUC=0.62) captures the change in gradient magnitude between the reference and test images and outperforms other features in detecting defocussed tampering. 
	\item Features $e_1, e_2$, and $k_1$, detect defocussed tampering better than other features. (It is not surprising that edge-based features are able to detect defocussed tampering better.)
	\item Feature $e_2$ outperforms other features in detecting moved tampering. 	 
	\item Overall covered tampering is detected better than defocussed and moved tampering. More features are able to detect defocussed tampering when compared to covered and moved tampering.
	\item We apply a constant threshold and hence the accuracies reported are not reflective of the upper limit on the detection ability of the feature. Adopting an adaptive thresholding mechanism can improve the detection accuracy for each feature. Through this analysis, our intent is not to maximize detection accuracy but rather compare each feature’s to detect tampering.
\end{itemize}

Performance at the optimal threshold: Optimal threshold is defined as the point on the ROC curve where the difference between TPR and FPR is maximum. We compute the performance of each feature on four groups of the tampered dataset, three containing exclusively one type of tampering and the fourth containing all types of tampering. The fourth dataset does not distinguish amongst the three tamperings and labels images as either tampered or normal. The performance on this dataset can help us ascertain the feature’s capability to detect tampering in general. The performance for covered, defocussed, moved, and unified tampering detection is summarized in Table~\ref{tab:performance_optimal} (Top Left),~\ref{tab:performance_optimal} (Top Right) ,~\ref{tab:performance_optimal} (Bottom Left) and~\ref{tab:performance_optimal} (Bottom Right), respectively.

\begin{table*}[]
	\center
	\begin{tabular}{|l|l|l|l|l|l|l|}
		\hline
		File & tn & fp & fn & tp & Accuracy & $f_1$ score\\
		\hline
		
		$b_1$&165061&26915&17109&4671&0.794&0.175\\
		$b_2$&177187&14789&13801&7979&0.866&0.358\\
		$b_3$&168806&23069&16677&5103&0.813&0.204\\
		$b_4$&164445&27531&14452&7328&0.803&0.258\\
		$e_1$&2639&189237&12&21768&0.114&0.187\\
		$e_2$&888&190988&0&21780&0.106&0.185\\
		$e_3$&2639&189237&12&21768&0.114&0.187\\
		$e_4$&121058&70818&7785&13995&0.632&0.262\\
		$k_1$&191876&0&21780&0&0.898&-\\
		$k_2$&3&191873&0&21780&0.101&0.185\\
		\hline
	\end{tabular}
	\begin{tabular}{|l|l|l|l|l|l|l|}
		\hline
		File & tn & fp & fn & tp & Accuracy & $f_1$ score\\
		\hline
		
		$b_1$&30913&161063&1545&20235&0.239&0.199\\
		$b_2$&29563&162413&1274&20506&0.234&0.200\\
		$b_3$&37663&154212&1878&19902&0.269&0.203\\
		$b_4$&48199&143777&2195&19585&0.317&0.211\\
		$e_1$&132640&59236&10972&10808&0.671&0.235\\
		$e_2$&150946&40930&11562&10218&0.754&0.280\\
		$e_3$&132640&59236&10972&10808&0.671&0.235\\
		$e_4$&191876&0&21780&0&0.898&0.0\\
		$k_1$&89556&102320&7415&14365&0.486&0.207\\
		$k_2$&109452&82424&10267&11513&0.566&0.198\\
		\hline
	\end{tabular}

	\begin{tabular}{|l|l|l|l|l|l|l|}
	
	\end{tabular}

	\begin{tabular}{|l|l|l|l|l|l|l|}
		\hline
		File & tn & fp & fn & tp & Accuracy & $f_1$ score\\
		\hline
		
		$b_1$&171707&20269&18774&3006&0.817&0.133\\
		$b_2$&52806&139170&4530&17250&0.327&0.193\\
		$b_3$&153428&38447&16335&5445&0.743&0.165\\
		$b_4$&38937&153039&3996&17784&0.265&0.184\\
		$e_1$&13334&178542&536&21244&0.161&0.191\\
		$e_2$&41363&150513&3004&18776&0.281&0.196\\
		$e_3$&13334&178542&536&21244&0.161&0.191\\
		$e_4$&95083&96793&8576&13204&0.506&0.200\\
		$k_1$&5428&186448&8&21772&0.127&0.189\\
		$k_2$&13738&178138&1144&20636&0.160&0.187\\
		\hline
	\end{tabular}
	\begin{tabular}{|l|l|l|l|l|l|l|}
		\hline
		File & tn & fp & fn & tp & Accuracy & $f_1$ score\\
		\hline
		
		$b_1$&164889&27087&54317&11023&0.683&0.213\\
		$b_2$&112537&79439&32832&32508&0.563&0.366\\
		$b_3$&154036&37839&49476&15864&0.660&0.266\\
		$b_4$'&48146&143830&10263&55077&0.401&0.416\\
		$e_1$&2639&189237&12&65328&0.264&0.408\\
		$e_2$&150209&41667&46963&18377&0.655&0.293\\
		$e_3$&2639&189237&12&65328&0.264&0.408\\
		$e_4$&103090&88786&29524&35816&0.540&0.377\\
		$k_1$&191876&0&65340&0&0.745&-\\
		$k_2$&6&191870&1&65339&0.254&0.405\\
		\hline
	\end{tabular}
	\caption{Performance at optimal threshold: (Top Left) Covered, (Top Right) Defocussed, (Bottom Left) Moved, and (Bottom Right) Unified tampering detection; . \label{tab:performance_optimal}}
\end{table*}

\begin{itemize}
	\item At the optimal threshold, $k_1$ fails to detect covered tampering, and labels all the images as normal. Table~\ref{tab:performance_optimal} (Top Left) shows a high accuracy value for $k_1$, as the ratio of normal images to covered images is large. As it fails to detect any covered images, $f_1$-score is not defined.
	\item Feature $b_2$ detects covered tampering with the highest accuracy and $f_1$-score.
	\item Features $b_1$, $b_2$, $b_3$, $b_4$, and $e_4$ display a capability to detect covered tampering $e_4$ produces a large number of true positives compared to other background based methods, however, it also generates a large number of false positives compared to the background based methods. 
	\item Feature $b_2$ generates the least number of false positives while detecting $35\%$ of the covered tampering. 
	\item Overall, feature based on background detected covered tampering better than feature based on edges and keypoints. 
	\item As shown in Table~\ref{tab:performance_optimal} (Top Right), feature $e_4$ fails to detect defocussed tampering completely and labels all the images as normal. It produces the highest accuracy for the same reason as $k_1$ in Table~\ref{tab:performance_optimal} (Top Left).
	\item Features based on background produced a large number of false positives when detecting defocussed tampering. 
	\item Feature $e_2$ produces the lowest false positives and is able to detect $45\%$ of the defocussed tampering. 
	\item In general, the false positives, generated while detecting defocussed is much larger than while detecting covered tampering. This is because the deviation in the residual due to defocussed tampering is smaller compared to covered tampering (See Table~\ref{tab:pred_srmse} (Right)).
	\item Excluding $e_4$, features based on the edge are capable of detecting defocussed better than those based on background.
	\item Features $b_1$, $b_3$ are capable of detecting moved tampering. $b_1$ is capable of detecting $13\%$ of moved tampering, and $b_3$ is capable of detecting $25\%$ of moved tampering. However, $b_3$ produces twice as many false positives as $b_1$.
	\item Features $e_4$ is capable of detecting, $60\%$ of moved tampering, but produces $4.5$ times more false positives than $b_1$.
	\item Under a unified tampering detection framework (Table~\ref{tab:performance_optimal} (Bottom Right)), $b_1$, $b_3$, $e_2$ are the better choices for tampering detection. Feature $b_1$ produces the least number of false positives while detecting $20\%$ tampering, followed by $b_3$ and $e_2$.
\end{itemize}

\vspace{-0.25cm}
\section{Conclusion}
We cast the problem of tampering detection as a subproblem of change detection. We have reviewed the corpus of published research in this area, with a focus on various feature types used for tampering detection. We have analyzed ten features, four based on background ($\{b_1, b_2, b_3, b_4\}$), four based on edges ($\{e_1, e_2, e_3, e_4\}$), and two based on keypoints ($\{k_1, k_2\}$). We formulate tampering detection as a time-series analysis problem and perform feature type analysis. We have ascertained that the complexity of modeling time series based on keypoint features is lower compared to the background and edge-based features. Feature $b_3$ is more robust to abrupt changes in videos. While $e_2$ appears to be the least robust feature, it has the highest capability to capture deviations due to tampering. Features $b_2$, $b_4$, and $e_4$ are capable of detecting covered tampering, features $e_1$, $e_2$, and $k_1$ are better are detecting defocussed tampering, and $e_2$ outperforms the other features at detecting moved tampering. Results suggest that the features are better at detecting covered tampering compared to defocussed and moved tampering, and features tend to produce a large number of false positives while detecting defocussed tampering. Furthermore, features based on background detected covered tampering better than those based on edges, and keypoints. We believe that existing hand-crafted features and decision-making mechanisms may be insufficient to address the complexity of the variations that occur due to tampering. There is much need to explore learned features, in addition to applying non-linear decision-making methods to achieve a robust and viable solution to tampering detection. 
\vspace{-0.3cm}
%
\bibliographystyle{IEEEtran}
\bibliography{IEEEexample,main}

\end{document}